\documentclass{article}

    \PassOptionsToPackage{numbers, compress}{natbib}
\usepackage[preprint]{neurips_2021}

\usepackage[utf8]{inputenc} % allow utf-8 input
\usepackage[T1]{fontenc}    % use 8-bit T1 fonts
\usepackage[colorlinks]{hyperref}       % hyperlinks
\usepackage{url}            % simple URL typesetting
\usepackage{booktabs}       % professional-quality tables
\usepackage{amsfonts}       % blackboard math symbols
\usepackage{nicefrac}       % compact symbols for 1/2, etc.
\usepackage{microtype}      % microtypography
\usepackage{xcolor}         % colors

\usepackage{graphicx}
\usepackage{subfigure}
\usepackage{float}
\usepackage{amsmath}

\title{Effective Eigendecomposition based Graph Adaptation for Heterophilic Networks}

\author{%
  Vijay Lingam\\
  Microsoft Research India\\
  \texttt{vijaylingam0810@gmail.com} \\
   \And
   Rahul Ragesh \\
   Microsoft Research India \\
   \texttt{rahulragesh@microsoft.com} \\
   \AND
   Arun Iyer \\
   Microsoft Research India \\
   \And
   Sundararajan Sellamanickam \\
   Microsoft Research India \\
}

\newcommand{\graph}{\mathcal{G}}
\newcommand{\adj}{\mathbf{A}}
\newcommand{\verts}{\mathcal{V}}
\newcommand{\edges}{\mathcal{E}}
\newcommand{\train}{T}

\newcommand{\feat}{\mathbf{X}}

\newcommand{\real}{\mathbb{R}}
\newcommand{\eigenU}{\mathbf{U}}
\newcommand{\eigenUc}{\mathbf{u}}
\newcommand{\adjS}{\mathbf{\Sigma}}
\newcommand{\adjSv}{\sigma}

\newcommand{\gcn}{\textsc{GCN}}
\newcommand{\sgcn}{\textsc{SGCN}}
\newcommand{\geomgcn}{\textsc{Geom-GCN}}
\newcommand{\supergat}{\textsc{SuperGAT}}
\newcommand{\hhgcn}{\textsc{H\textsubscript{2}GCN}}
\newcommand{\fagcn}{\textsc{FAGCN}}
\newcommand{\gprgnn}{\textsc{GPR-GNN}}
\newcommand{\eigennet}{\textsc{EigenNetwork}}
\newcommand{\eigenconcat}{\textsc{Eigen-ConcatNetwork}}
\newcommand{\eigeneigennet}{\textsc{Eigen-EigenNetwork}}

\newcommand{\regeigeneigenet}{\textsc{RegEigen-EigenNetwork}}

\newcommand{\appnp}{\textsc{APPNP}}

\begin{document}

\maketitle

\begin{abstract}
Graph Neural Networks (GNNs) exhibit excellent performance when graphs have strong homophily property, i.e. connected nodes have the same labels. However, they perform poorly on heterophilic graphs. Several approaches address the issue of heterophily by proposing models that adapt the graph by optimizing task-specific loss function using labelled data. These adaptations are made either via attention or by attenuating or enhancing various low-frequency/high-frequency signals, as needed for the task at hand. More recent approaches adapt the eigenvalues of the graph. One important interpretation of this adaptation is that these models select/weigh the eigenvectors of the graph. Based on this interpretation, we present an eigendecomposition based approach and propose \eigennet~models that improve the performance of GNNs on heterophilic graphs. Performance improvement is achieved by learning flexible graph adaptation functions that modulate the eigenvalues of the graph. Regularization of these functions via parameter sharing helps to improve the performance even more. Our approach achieves up to 11\% improvement in performance over the state-of-the-art methods on heterophilic graphs.
\end{abstract}

\section{Introduction}
Homophily~\citep{homophily} is a principle in sociology that suggests that similarity breeds connections in real life. In the context of semi-supervised classification, this implies that nodes with similar labels are likely to be connected. Several real-world networks exhibit homophily; for example, people on a social network connect based on similar interests. Several real-world networks exhibit the opposite as well, which is heterophily. For example, the Wikipedia page on homophily is linked to other pages from sociology and connected to various pages from mathematics, graph theory, and statistics. %Since Wikipedia is a large body of collective knowledge, its pages often have connections between several different areas.

Graph Neural Networks (GNNs)~\citep{gcn, graphsage, gat} leverage network information along with node features to improve their semi-supervised classification performance. GNNs are primarily dependent on network homophily to be able to give improved performance. For heterophilic networks, their performance can degrade significantly. Several approaches have been proposed in the literature to mitigate this degradation in performance in the presence of heterophily. \citet{geomgcn} aggregates both over the graph neighbourhood and the neighbours in the latent space. However, the neighbours still influence the self-embedding of the central node and could bring in noise. ~\citet{h2gcn} keeps the self-embedding separate from the neighbour embeddings during aggregation, while also similarly incorporating higher-order neighbour embeddings. ~\citet{supergat} proposed several simple attention models trained on an additional auxiliary task. 
% The second way to address heterophily is to model a label-label compatibility matrix explicitly. This matrix acts as a prior in updating the posterior belief in the label predictions. \citet{cpgnn} proposes to model the label compatibility matrix that reflects the heterophily in the graph and utilizes the model in a GNN. The more recent approaches directly adapt the low-frequency and high-frequency parts of the graph as needed by the model. 
\citet{fagcn} proposes to learn an attention mechanism that captures the proportion of low-frequency and high-frequency signals per edge. \citet{gprgnn} proposes an adaptive polynomial filter to pick up which low-frequency or high-frequency signals are helpful for the task. There have also been other approaches involving label-label compatibility matrix~\citep{cpgnn}.

For homophilic networks, existing models~\citep{appnp, gprgnn} already prove to be excellent. Our interest lies in heterophilic networks. We mainly focus on the class of methods that aim at modifying or adapting the graph to obtain better performance~\citep{supergat, fagcn, gprgnn} in heterophilic networks. The more recent among these approaches adjust the eigenvalues of the graph to learn improved representation. Another interpretation for this adaptation is that it is suppressing some eigenvectors while accentuating others. We use this insight and propose simple yet effective methods of weighting the eigenvectors to improve task performance. We make the following contributions in this work:
\begin{enumerate}
    \item We present a simple eigendecomposition based approach and propose \textsc{EigenNetwork} models to learn flexible graph adaptation functions. We show that our models achieve significantly improved performance where the graphs are heterophilic. 
    \item While most GNNs are aggregation-based models, we propose a simple and efficient concatenation model that is quite competitive with neighborhood aggregation models on several datasets.
    \item Finally, we propose a weight-tying based regularization to learn better adaptation functions to avoid any possible over-fitting, when learning from limited data.
    \item We conduct extensive experimentation and our approach achieves up to 11\% improvement in performance over the state-of-the-art methods on heterophilic graphs.
\end{enumerate}

In the following sections, we discuss related works (Section~\ref{sec:relatedwork}). We motivate our work with the recently proposed model~\citep{gprgnn} in Section~\ref{sec:preliminaries}. We give details about our proposed approach in Section~\ref{sec:proposedapproach}. Finally, we give our experiment results, ablative studies and conclusion in Sections~\ref{sec:experiments} and~\ref{sec:conclusion}.

\section{Related Works}
\label{sec:relatedwork}
% In recent times, Graph Neural Networks (GNNs) have become an increasingly popular method for semi-supervised classification with graphs. \citet{spectralnetwork} proposed an early version of the GNN where they generalized the idea of Convolutional Neural Networks (CNNs) to graphs by defining convolutions over the graph in the Fourier domain, but it had non-spatially localized graph filters. \citet{chebgnn} proposed to approximate the filters using Chebyshev polynomial, which gave a spatially localized filter. \textsc{GCN}~\citep{gcn} further approximated the Chebyshev polynomial to the first-order neighbourhood, providing a faster and a much simpler variant. In~\gcn, the convolution operation reduces to aggregating features over the neighbourhood. Improving the aggregation mechanism~\citep{graphsage, gat} and incorporating random walk information~\citep{ngcn, mixhop, deeperinsights} gave further improvements in these models. However, all these models suffered from one common issue: increasing receptive field led to over smoothing features, thereby causing model performance to degrade. To circumvent this problem, \appnp~\citep{appnp} proposed an approach derived from personalized Pagerank, thereby enabling it to incorporate long-range information without degradation in performance.

In recent times, Graph Neural Networks (GNNs) have become an increasingly popular method for semi-supervised classification with graphs. \citet{spectralnetwork} set the stage for early GNN models followed by various modifications~\citep{chebgnn, gcn}. \textsc{GCN}~\citep{gcn} provided the fastest and simplest variant, where the convolution operation reduces to aggregating features over the neighbourhood. Improving the aggregation mechanism~\citep{graphsage, gat} and incorporating random walk information~\citep{ngcn, mixhop, deeperinsights} gave further improvements in these models, but they still suffered from over smoothing. To circumvent this problem, \appnp~\citep{appnp} proposed an approach derived from personalized Pagerank.

Most of the development in the GNNs were for homophilic graphs, and they performed poorly in heterophily setting. One of the early works to address heterophily in GNNs was \geomgcn~\citep{geomgcn}. They identified two key weaknesses in GNNs in the context of heterophily. Firstly, since the aggregation over the neighbourhood is permutation-invariant, it is difficult to identify which neighbours contribute positively and negatively to the final performance. Secondly, long-range information is difficult to aggregate. To mitigate these issues, they proposed aggregating over two sets of the neighbourhood - one from the graph and the other inferred in the latent space. \hhgcn~\citep{h2gcn} proposed to separate the self-embeddings from neighbour embeddings. To avoid mixing of information, they concatenate self-embeddings and neighbour embeddings instead of aggregating them. Higher-order neighbourhood embeddings are similarly combined to capture long-range information. 

Recent approaches address the weaknesses by adapting the graph itself. \supergat~\citep{supergat} gave several simple attention models trained on the classification and an additional auxiliary task. They suggest that these attention models can improve model performance across several graphs with varying homophily scores. \fagcn~\citep{fagcn} uses the attention mechanism and learns the weight of an edge as the difference in the proportion of low-frequency and high-frequency signals. They empirically show that negative edge-weights identify edges that connect nodes with different labels. \gprgnn~\citep{gprgnn} takes the idea proposed in \appnp~and generalizes the Pagerank model that works well for graphs with varying homophily scores. Our proposed approach is closely related to these methods that adapt graph for the task at hand. Mainly, we take inspiration from the \gprgnn. We show that \gprgnn~effectively adapts the eigenvalues of the graph for the desired task via the polynomial with learnable coefficients~(Section~\ref{sec:proposedapproach}). We propose to replace this polynomial with a \textit{graph adaptation} function, which allows our eigengraph based network model (\eigennet) to learn any sharp changes in the importance of eigenvectors. It enables our model to give better performance on several datasets. Additionally, we observe that instead of aggregating features over the graph, we can get competitive performance with simpler models in some datasets if we \textit{concatenate} our adapted graph with the features.

In the following section, we give a brief overview of the problem setting and \gprgnn~model focussing on the key elements and ideas to motivate our work.

\section{Problem Setup and Motivation}
\label{sec:preliminaries}
% \begin{table}[ht]
% \begin{tabular}{ll}
% $n$ & number of nodes \\
% $d$ & number of features \\
% $\adj \in \{0, 1\}^{n \times n}$ & Adjacency Matrix \\
% $\eye$ & Identity Matrix\\
% $\adji = \adj + \eye$ & Adjacency + Identity \\
% $\dadji$ & Degree matrix associated with $\adji$ \\
% $\feat \in \real^{n \times d}$ & Feature Matrix \\
% $\adjU, \adjV$ & Eigenvectors of $\adji$ \\
% $\lambda_i$ & Singular values of $\adji$, $\lambda_i \geq \lambda_j, \;\; i < j$ \\
% $\featU, \featV$ & Eigenvectors of $\feat$ \\
% $\sigma_i$ & Singular values of $\feat$, $\sigma_i \geq \sigma_j, \;\; i < j$ \\
% $h$ & Filter Function 
% \end{tabular}
% \end{table}

We focus on the problem of semi-supervised node classification on a simple graph $\graph = (\verts, \edges)$, where $\verts$ is the set of vertices and $\edges$ is the set of edges. Let $\adj \in \{0, 1\}^{n \times n}$ be the adjacency matrix associated with $\graph$, where $n = |\verts|$ is the number of nodes. Let $\mathcal{Y}$ be the set of all possible class labels. Let $\feat \in \real^{n \times d}$ be the $d$-dimensional feature matrix for all the nodes in the graph. Given a training set of nodes $\train \subset \verts$ whose labels are known, along with $\adj$ and $\feat$, our goal is to predict the labels of the remaining nodes. The proportion of edges that connect two nodes with the same labels in a graph is called the homophily score of the graph. In our problem, we are particularly concerned with graphs that exhibit low homophily scores. In the next sub-section, we provide background material on the GPR-GNN modelling method and its approach to graph adaptation.

\subsection{\gprgnn~Model}  
The \gprgnn~\citep{gprgnn} model consists of two core components: (a) a non-linear network that transforms raw feature input $\feat$: ${\bf Z}^{(0)} = f(\feat;{\mathbf W})$ and (b) a generalized page ranking (GPR) component, ${\mathbf G}$, that essentially aggregates  the transformed output ${\mathbf Z}$ recursively as: ${\bf Z}^{(k)} = \adj{\bf Z}^{(k-1)}, k = 1, \ldots, K$. Notice that there is no nonlinear operation involved after each aggregation step over $k$. Therefore, the functionality of the GPR component can be written using an operator ${\mathbf G}$ defined as: ${\mathbf G} = \sum_{k=0}^K \alpha_k {\mathbf A}^k$
%\begin{equation}
%{\mathbf G} = \sum_{k=0}^K \alpha_k {\mathbf A}^k
%\label{eqn:GPRGNN}
%\end{equation}
and we obtain aggregated node embedding by applying ${\mathbf G}$ on the nonlinear network output:  ${\mathbf S} = {\mathbf G} {\mathbf Z}^{(0)}$. 
%Then the class probability distribution of nodes are computed as: ${\mathbf P} = \textsc{Softmax}({\mathbf S} {\mathbf{W}_c)}$ where ${\mathbf W}_c$ denote the classifier model weights. Model parameters $({\bf W}, {\mathbf W}_c, {\mathbf \alpha}$) are learned using labeled data by optimizing standard cross-entropy loss function. 
Using eigendecomposition, ${\adj} = \eigenU \adjS \eigenU^T$, \citet{gprgnn} presented an interpretation that the GPR component essentially performs a graph filtering operation: ${\mathbf G} = \eigenU h_K(\adjS) \eigenU^T$ where $h_K(\adjS)$ is a polynomial graph filter applied element-wise and $h_K(\adjSv_i) = \sum_{k=0}^K \alpha_k \adjSv^k_i$ where $\adjSv_i$ is the $i^{th}$ eigen value. As explained in~\citet{gprgnn}, learning filter coefficients (i.e.,$\alpha$) help to get improved performance. Since the coefficients can take negative values the \gprgnn~model is able to capture high-frequency components of the graph signals, enabling the model to achieve improved performance on heterophilic graphs.
%Using eigendecomposition of $\adj$, \citet{gprgnn} presented an interpretation that the GPR component essentially performs a graph filtering operation, as follows. Let ${\adj} = \eigenU \adjS \eigenU^T$ be the eigendecomposition. On substitution in (\ref{eqn:GPRGNN}), we get: ${\mathbf G} = \eigenU h_K(\adjS) \eigenU^T$ where $h_K(\adjS)$ is a polynomial graph filter applied element-wise and $h_K(\adjSv_i) = \sum_{k=0}^K \alpha_k \adjSv^k_i$ where $\adjSv_i$ is the $i^{th}$ eigen value. As explained in~\citet{gprgnn}, learning filter coefficients (i.e.,$\alpha$) help to get improved performance. Since the coefficients can take negative values the \gprgnn~model is able to capture high-frequency components of the graph signals, enabling the model to achieve improved performance on heterophilic graphs. 
In the following sub-section, we analyze the importance of eigenvectors in heterophilic graphs.

\subsection{Eigenvector Analysis}
\label{sec:eigenvectoranalysis}
\begin{figure}%
    \centering
    \subfigure[Crocodile]{
        \label{fig: crocodile_partial}
        \includegraphics[width=0.3\textwidth]{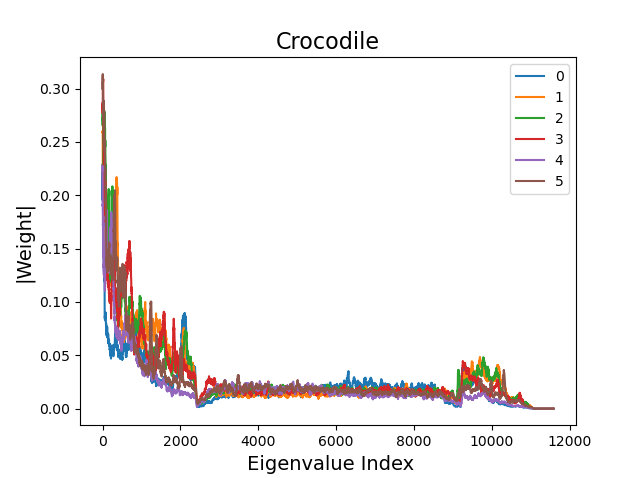}
    }
    \subfigure[Chameleon]{
        \label{fig:chameleon_partial}
        \includegraphics[width=0.3\textwidth]{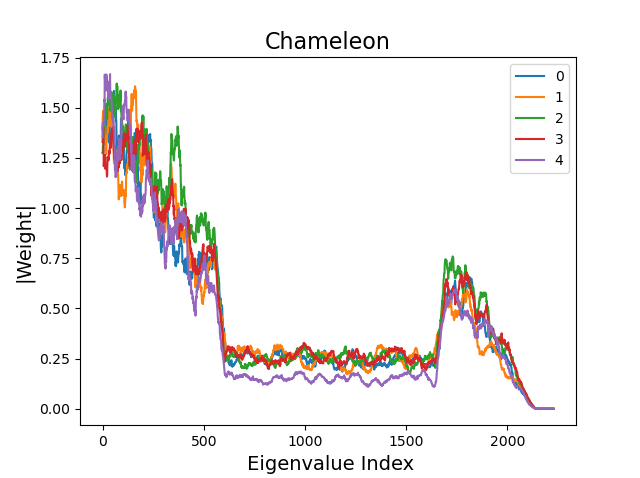}
    }
    \subfigure[Squirrel]{
        \label{fig:wisconsin_partial}
        \includegraphics[width=0.3\textwidth]{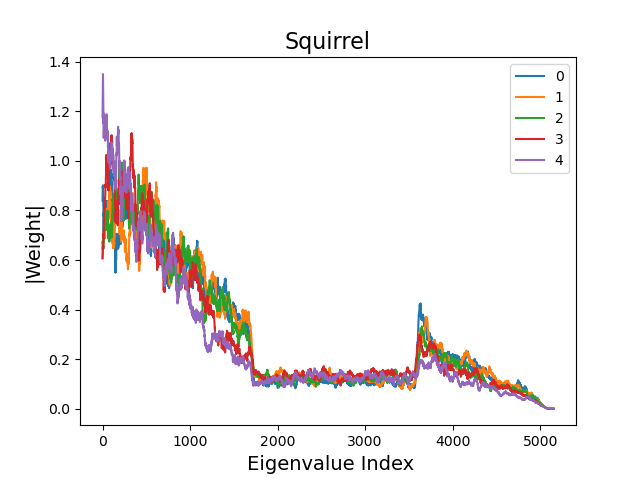}
    }
    % \\
    % \subfigure[Crocodile-FL]{
    %     \label{fig:crocodile_full}%
    %     \includegraphics[width=0.3\textwidth]{images/weightPlots/crocodile_fl.png}
    % }
    % \subfigure[Chameleon-FL]{
    %     \label{fig:chameleon_full}%
    %     \includegraphics[width=0.3\textwidth]{images/weightPlots/chameleon_fl.png}
    % }
    % \subfigure[Wisconsin-FL]{
    %     \label{fig:wisconsin_full}
    %     \includegraphics[width=0.3\textwidth]{images/weightPlots/wisconsin_fl.png}
    % }    
    \caption{Weight Plots}
    \label{fig:weightPlots}
\end{figure}

We conduct the following experiment to study why adapting eigenvalues is important in heterophilic settings empirically. We consider the eigendecomposition of given graph as ${\adj} = \eigenU \adjS \eigenU^T$ and we assign the node features as $\eigenU \adjS^{\frac{1}{2}}$. The columns in this feature matrix are essentially the eigenvectors. We construct new training data with these features using the known labels from training set~$\train$. We train a Logistic Regression model with this data. Figure~\ref{fig:weightPlots} plots the per-class (indicated by different colors) weights learnt by the model. The x-axis is the index of the eigenvalues sorted in descending order. The y-axis denotes the weight assigned to the corresponding eigenvector. These weights give the importance of eigenvectors. These plots show the spread of useful signals across the spectrum. For instance, the Chameleon and Crocodile dataset exhibit a dumbbell-like distribution. Both low-frequency and high-frequency components have high (absolute) weights suggesting that selecting/weighing eigenvectors is the key. Correctly selecting/weighing them will give us a good performance on heterophilic graphs. In the next section, we give our proposed approach for adapting the graph based on the analysis presented here.

\section{Proposed Approach: Eigen Network Models}
\label{sec:proposedapproach}
In this section, we present an alternate interpretation of the \gprgnn~model and suggest a simple eigendecomposition based graph adaptation approach. We present several model variants, each one of them is motivated by considering different aspects of the problem.
\subsection{Eigen Network Model} 
We start by observing closely the GPR component output given by: 
\begin{equation}
    {\mathbf S} = \eigenU h_K(\adjS) \eigenU^T {\mathbf Z}^{(0)}.
\label{eqn:GPRGNN2}    
\end{equation}
Our first observation is that learning filter coefficients is equivalent to learning a \textit{new} graph, ${\tilde \adj(\eigenU, \adjS;\alpha})$ which is dependent on the fixed set of eigenvectors and eigenvalues, but, parameterised using $\alpha$. Therefore, the \gprgnn~model may be interpreted as adapting the original adjacency matrix $\adj$. Next, as noted in the previous section, using the structures present in eigendecomposition and polynomial function, we can expand (\ref{eqn:GPRGNN2}) by unrolling over eigenvalues and interchanging the summation as:
\begin{equation}
    {\mathbf S} = \sum_{j=1}^n \eigenUc_j h_K(\adjSv_j;\alpha) \eigenUc^T_j {\mathbf Z}^{(0)}.
\label{eqn:GPRGNN3}    
\end{equation}
Since $h_K(\adjSv_j;\alpha)$ is dependent only on $(\adjSv_j;\alpha)$ our proposal is to replace the polynomial function with any general smooth function, $C(\sigma_i;\alpha)$, which need not be a polynomial and we call $C(\cdot)$ as \textit{graph adaptation} function. We discuss several choices of such functions shortly. We rewrite (\ref{eqn:GPRGNN3}) in matrix form, after substituting the graph adaptation matrix and leaving out the input embedding (${\mathbf Z}^{(0)}$) as:  
\begin{equation}
    {\mathbf E} = \eigenU C(\adjS;\alpha) \eigenU^T
\label{eqn:eigenn}
\end{equation}
%where $\eigenU_x = \eigenU^T {\mathbf Z}^{(0)}$. 
 We refer to (\ref{eqn:eigenn}) as \eigennet~as it involves eigenvectors and learnable graph adaptation function that is dependent on eigenvalues. This network forms the basis of our eigendecomposition based modeling approach. Note that (\ref{eqn:eigenn}) is essentially a single layer network.

{\bf Choice of Graph Adaptation Function.} We use graph adaptation function that is a non-negative function of eigenvalue. One general function is: $C(\adjSv;\alpha) = g_s(\alpha_1) \adjSv^{g_e(\alpha_2)}$ where we use subscripts to differentiate scaling and exponent activation functions. We find that $\textsc{ReLu}(\cdot)$ is useful for both scaling and exponentiation, and $\textsc{Sigmoid}(\cdot)$ is also useful for scaling purpose. Composition with two functions is quite flexible and helps to adapt for graphs having diverse eigenvalue decay rates. Note that we can recover original eigenvalues for suitable choices of $\alpha_1$ and $\alpha_2$. There may be other choices of graph adaptive functions that perform better. From a practical viewpoint, several functions can be evaluated with the best function selected using traditional hyperparameter optimization strategy. 

{\bf Regularization of Graph Adaptation Function.} While it is possible to use separate set of parameters for each eigenvalue, the number of model parameters can go up significantly, depending on the number of eigenvectors. We can mitigate this problem, for example, by using same parameter, $\alpha$ for several eigenvalues. Weight tying is a popular mechanism of regularization used in CNN~\citep{cnn}, statistical relational learning~\citep{srl}, Markov logic networks~\citep{mln}, probabilistic soft logic~\citep{psl}, language models~\citep{transformer} etc. In Figure~\ref{fig:weightPlots}, we observe that nearby eigenvalues tend to have similar importance. Based on this observation, we divide the sorted set of eigenvalues in fixed-length bins and assign one $\alpha$ variable to each bin. We call this model as \regeigeneigenet. This allows us to reduce the number of learnable parameters in the model and offer a simple but effective regularization.

{\bf Computational Aspect.} One main difficulty arises when the graph adaptation parameters are jointly learned with the nonlinear feature transformation and classifier model weights. This is because computing node  embedding (\ref{eqn:GPRGNN2}) using (\ref{eqn:eigenn}) to transform input features (i.e., ${\mathbf E} {\mathbf Z}^{(0)}$) involves dense matrix multiplications and is expensive. We can reduce the computational cost in several ways: (1) We use raw input features $\feat$. This helps to pre-compute projected features, $\eigenU_x = \eigenU^T \feat$. The other possibility is to use pre-trained feature embedding and use $\eigenU^T {\mathbf Z}^{(L)}$. (2) It also helps to reduce the dimension of input features whenever the dimension of raw features or pre-trained embedding is high. We provide more details in Section~\ref{sec:EigenEigen}. (3) Since we are adapting the eigenvalues we may not need a large number of eigenvectors to get good performance, as validated in our experiments. %For ease of discussion, we work with raw features. However, the following approach holds for pre-trained embedding as well. 

{\bf Remarks.} We observe that the learned graph is nothing but $\adj(\alpha) = \eigenU{\mathbf C}(\adjS;\alpha)\eigenU^T$ and is dense with all nodes in the graph are essentially used to learn node embedding. 
%Unlike $\adj^K$, the learned graph can have negative edges and they are optimized using labeled data. This capability to have negative edges enable our \eigennet~model to perform better on heterophily graphs, as we demonstrate in the experimental section. 
From the experimental study presented in Section~\ref{sec:eigenvectoranalysis} we see that the adaptation function required to improve performance can be quite complex for heterophily graphs. Therefore, GPR-GNN that uses polynomial functions to learn the adaptation function may not be able to adapt the graph effectively in some situations. On the other hand, our approach has the ability to learn more complex adaptation function and is expected to perform better. It is worth noting that the necessity of negative edges has been highlighted in~\citet{fagcn} and~\citet{gprgnn} using graph filtering concept. \citet{fagcn} use attention mechanism to learn negative edges. On the other hand, \citet{gprgnn} uses polynomial function with negative weights to obtain negative edges. In contrast, our approach to learn negative edges is quite different. It emerges from adapting eigengraphs ($\eigenUc_j\eigenUc_j^T$) starting with good initialization obtained from approximating the graph using eigendecomposition. Our experimental results show that the proposed approach is highly powerful and outperforms both \fagcn~and \gprgnn~on several heterophily datasets.    

\subsection{\label{sec:EigenEigen} Eigen-Eigen Network Model}
It is often useful to reduce dimension of raw features using principal component analysis. Let  ${\mathbf Q} \adjS_x {\mathbf Q}^T$ be the eigendecomposition of $\feat\feat^T$.  
%Let ${\mathbf X} = {\mathbf Q}_x {\mathbf \Sigma}_x {\tilde \mathbf Q}^T_x$ be the singular value decomposition of ${\mathbf X}$. Note that ${\mathbf X}$ is usually any general matrix. 
Using this decomposition, we define parameterised node embedding for $\feat$ as: ${\tilde \feat}(\beta) = {\mathbf Q} {\mathbf C}_x(\adjS_x;\beta)$. Upon substituting node embedding for raw features in the \eigennet~model (\ref{eqn:eigenn}), we get: 
\begin{equation}
    {\mathbf S} = \eigenU {\mathbf C}(\adjS;\alpha) \eigenU_{\tilde x} {\mathbf C_x}(\adjS_x;\beta)
\label{eqn:eigeignn}
\end{equation}
where $\eigenU_{\tilde x} = \eigenU^T {\mathbf Q}$. We refer (\ref{eqn:eigeignn}) as \eigeneigennet~as it involves eigenvectors of both $\adj$ and $\feat$, and, involves learning weights for eigenvectors. Making use of the fact that ${\mathbf C}(\cdot)$ and ${\mathbf C_x}(\cdot)$ are diagonal, we rewrite (\ref{eqn:eigeignn}) as:   
\begin{equation}
    {\mathbf S} = \eigenU {\mathbf F}(\cdot;\alpha,\beta)
\label{eqn:eigeignn2}
\end{equation}
where ${\mathbf F}(\cdot;\alpha,\beta) = {\mathbf U}_{\tilde x} \bigodot {\mathbf c}(\adjS;\alpha) {\mathbf c}^T_x(\adjS_x;\beta)$ and $\bigodot$ denote element-wise product operation. ${\mathbf c}(\cdot)$ and ${\mathbf c}_x(\cdot)$ denote vectors of diagonal entries. Model learning involves learning $({\mathbf W}_c,\alpha, \beta)$ using labeled data by optimizing cross-entropy loss function. In this paper, we are primarily interested in adapting the graph. Therefore, we optimize only $({\mathbf W}_c,\alpha)$ in (\ref{eqn:eigeignn2}), keeping ${\mathbf C}_x$ fixed to eigen values. We leave conducting an experimental study with $\beta$ optimization as a future work. 

\subsection{Eigen-Concat Models} 
We suggest a simple alternate modeling approach that works quite well for heterophilic graphs. The motivation for this approach is two-fold. While conducting the empirical study discussed in Section~\ref{sec:eigenvectoranalysis}, we found that eigenvectors of ${\adj}$ alone can give good performance for heterophily graphs and neighborhood aggregation using traditional methods only degrade the performance. 
%Likewise, node features alone are sufficient in some situation to achieve very good performance without neighborhood aggregation. Furthermore, 
Furthermore, difficulties arise when ${\adj}$ and ${\mathbf X}$ are \textit{incompatible} in the sense that neighborhood aggregation degrades the performance due to violation of assumptions made. Though graph adaptation methods mitigate the effect of any violation, they still operate within the field of improving neighborhood aggregation. Therefore, it may be difficult to improve beyond some limit with the neighborhood aggregation restriction. Also, it may only add more computational burden. In this context, we explored the approach of concatenating node features (${\mathbf X})$ (or transformed features, ${\mathbf Z}$) and fixed or adapted eigen vectors of $\adj$, and learning a classifier model. Note that learning the adaptation function using only $\eigenU {\mathbf C}(\adjS;\alpha)$ is possible. Since the features are decoupled now, there is significant reduction in computational cost. Therefore, the Eigen-Concat model is faster to train. We found this simple approach to be competitive in several heterophily benchmark datasets. Therefore, it is a simple but important baseline to include in any work that aim at improving performance on heterophily graphs.     

%[{\bf To be written}]
%Suggest as an alternative to aggregation model with some motivation from (a) \lgcn/\hetegcn~work and (b) how disagreement in $\adj$ and $\feat$~can hurt each other, and say concat model may be beneficial in some situation as disagreements may be easy to resolve in concat model by giving feature importance weight. 

\subsection{Model Training and Complexity} 
\label{sec:complexity}
Let $p(\cdot;{\mathbf W},\alpha)$ and ${\bf y}$ denote the model predicted probability vector and 1-hot binary representation of true class label. We use the following standard cross-entropy based loss function: 
%\begin{equation}
    $${\mathcal L}({\mathbf W},\alpha) = -\sum_{i \in \train} \sum_{j = 1}^m y_{i,j} \log p_{i,j}(\cdot;{\mathbf W},\alpha)$$
%\label{eqn:objfn}
%\end{equation}
where $\train$ and $m$ denote the set of labeled examples and number of classes respectively, and indices $i$ and $j$ index examples and classes respectively. The model predicted probabilities are computed using \textsc{Softmax} function on classifier model scores, ${\mathbf S} {\mathbf W}_C$. %Note that the node embedding matrix (${\mathbf S})$ and its parameterization vary across models, and (\ref{eqn:objfn}) is an instance with parameters indicated for our models. 

Let $J_a$ and $J_x$ denote the number of eigen components used in our model. Assume that the graph adaptive function is parameterized with $J_g$ parameters for each of the $J_x$ component. Then, the number of model parameters needed in the \eigennet~model (\ref{eqn:eigeignn2}) is $J_g J_a + J_x$. As explained earlier, we do not optimize $\beta$ in our graph adaptation experiments. Given ${\mathbf U}_{\tilde x}$, the cost of computing ${\mathbf F}(\cdot)$ in (\ref{eqn:eigeignn2}) is $J_a J_d$. Likewise, given ${\mathbf U}$, the embedding computing cost per node is $J_a J_x$. Since (\ref{eqn:eigeignn2}) requires fresh computation whenever ${\mathbf F}(\cdot)$ is updated, our method is computationally expensive compared to the \gprgnn~method.

\section{Experiments}
\label{sec:experiments}
We validate our proposed models by comparing against several baselines and state-of-the-art heterophily graph networks on node classification task. In Section~\ref{baselines} we describe the baseline models and hyper-parameters tuning setup. In Section~\ref{impl_details} we describe our proposed models implementation details. In Section~\ref{hetero_exps}, we present our main experimental results on Heterophilic datasets. Although our key focus is on Heterophilic datasets, we present results on Homophilic datasets in Section~\ref{homo_exps}. Finally, in Section~\ref{analysis}, we present analysis and ablative studies.

\subsection{Baselines}
\label{baselines}
We provide the methods in comparison along with the hyper-parameters ranges for each model. For all the models, we sweep the common hyper-parameters in same ranges. Learning rate is swept over [0.001, 0.003, 0.005, 0.008, 0.01], dropout over [0.2, 0.3, 0.4, 0.5, 0.6, 0.7, 0.8], weight decay over [1e-4, 5e-4, 1e-3, 5e-3, 1e-2, 5e-2, 1e-1], and hidden dimensions over [16, 32, 64]. For model specific hyper-parameters, we tune over author prescribed ranges. We use undirected graphs with symmetric normalization for all graph networks in comparison. For all models, test accuracy is reported for the configuration that achieves the highest validation accuracy. We report standard deviation wherever applicable. 

\textbf{LR and MLP:} We trained Logistic Regression classifier and Multi Layer Perceptron on the given node features. For MLP, we limit the number of hidden layers to one. 

\textbf{\sgcn:} \sgcn~\citep{sgcn} is a spectral method that models a low pass filter and uses a linear classifier. The number of layers in \sgcn~is treated as a hyper-parameter and swept over [1, 2].  

\textbf{\supergat:} \supergat~\citep{supergat} is an improved graph attention model designed to also work with noisy graphs. \supergat~ employs a link-prediction based self-supervised task to learn attention on edges. As suggested by the authors, on datasets with homophily levels lower than 0.2 we use \supergat\textsubscript{SD}. For other datasets, we use \supergat\textsubscript{MX}. We rely on authors code\footnote{https://github.com/dongkwan-kim/SuperGAT} for our experiments.

\textbf{\geomgcn:} \geomgcn~\citep{geomgcn} proposes a geometric aggregation scheme that can capture structural information of nodes in neighborhoods and also capture long range dependencies. We quote author reported numbers for Geom-GCN. We could not run Geom-GCN on other benchmark datasets because of the unavailability of a pre-processing function that is not publicly available.

\textbf{\hhgcn:} \hhgcn~\citep{h2gcn} proposes an architecture, specially for heterophilic settings, that incorporates three design choices: i) ego and neighbor-embedding separation, higher-order neighborhoods, and combining intermediate representations.  We quote author reported numbers where available, and sweep over author prescribed hyper-parameters for reporting results on the rest datasets. We rely on author's code\footnote{https://github.com/GemsLab/H2GCN} for our experiments.

\textbf{\fagcn:} \fagcn~\citep{fagcn} adaptively aggregates different low-frequency and high-frequency signals from neighbors belonging to same and different classes to learn better node representations. We rely on author's code\footnote{https://github.com/bdy9527/FAGCN} for our experiments.

\textbf{\appnp:} \appnp~\citep{appnp} is an improved message propagation scheme derived from personalized PageRank. \appnp's addition of probability of teleporting back to root node permits it to use more propagation steps without oversmoothing. We use \gprgnn's~ implementation of \appnp~for our experiments.

\textbf{\gprgnn:} \gprgnn~\citep{gprgnn} adaptively learns weights to jointly optimize node representations and the level of information to be extracted from graph topology. We rely on author's code\footnote{https://github.com/jianhao2016/GPRGNN} for our experiments.

\subsection{Implementation Details}
\label{impl_details}

In this subsection, we present several important points that are useful for practical implementation of our proposed methods and other experiments related details. The eigendecomposition approach is based on adaptation of eigen graphs constructed using eigen components. Following~\cite{gcn}, we use a symmetric normalized version ($\tilde{\mathbf A}$) of adjacency matrix ${\mathbf A}$ with self-loops: $\tilde{\mathbf A} = {\tilde D}^{-\frac{1}{2}} ({\mathbf A} + {\mathbf I}) {\tilde D}^{-\frac{1}{2}}$ where ${\tilde D}_{ii} = 1 + D_{ii}$, $D_{ii} = \sum_{j} A_{ij}$ and ${\tilde D}_{ij} = 0, i \ne j$. We work with eigen matrix and eigen values of $\tilde {\mathbf A}$. From a practical viewpoint, it is difficult to work with all eigen components for two reasons: (a) the method becomes infeasible for large graphs and (b) in many applications, we do not need all eigen components and a fairly small to moderate number of components are sufficient to get excellent performance (See Figure~\ref{fig:varyEigPlots}). Noting the relation between singular and eigen vectors/values of a symmetric matrix~\citep{golub}, we use top-$d$ singular vectors/values of ${\tilde{\mathbf A}}$ in all our experiments. $\eigenU_d$ and $\adjS_d$ consists of top-$d$ singular vectors and singular values respectively. We used $d=\min(n,2048)$ unless otherwise specified, where $n$ denotes the numbers of nodes. We provide additional details on our proposed models below.

% \textbf{\eigennet\ Models:} We experimented with two different adaptation functions (i) $\textsc{Sigmoid}(\cdot)$ as $g_s$ and $\textsc{ReLu}(\cdot)$ as $g_e$. Each eigenvalue is parameterized by an individual $g_s$, but $g_e$ is shared. (ii) Both $g_{s}$ and $g_{e}$ uses $\textsc{ReLU}$. They are learned for each eigenvalue. Additionally, we observed that scaling 
% $\eigenU$~ by a constant factor gave improved results for few datasets. We tried two different scale factors [1,10]. For \tiedeigen~models, there is an additional $l_1$ weight regularization applied to $g_s$ and swept in the range [1e-3, 1e3] in logarithmic steps.

% \subsection{Eigen Network Models}
\textbf{\eigennet}. Recall that \eigennet~ uses only the graph, ${\tilde{\mathbf A}}$, to learn embedding and subsequently, the classifier model. Specifically, the score matrix is to be computed as: ${\mathbf S} = \eigenU_d C(\adjS_d;\alpha) \eigenU_d^T {\mathbf W}_c$ where $C(\cdot)$ is a learnable adaptation function. However, learning this model is quite expensive for large graphs. Therefore, we simplify our model by substitution: ${\mathbf W}'_c$ = $\eigenU_d^{T}{\mathbf W}_c$ and can learn ${\mathbf W}'_c$ directly for linear models. This helps to avoid expensive matrix multiplication with $\eigenU^T$. Using the same idea, we simplify our nonlinear model and learn \eigennet~ embedding as: ${\mathbf E} = \eigenU_d C(\adjS_d;\alpha)$ that is fed to a non-linear network; here, $d$ denotes the number of eigen components. Note that this model uses only the topological/graph features and do not leverage the available node features. 
\par 
\textbf{\eigeneigennet}. 
\eigeneigennet~ is defined as ${\mathbf E} = \eigenU_d {\mathbf F}(\cdot;\alpha,\beta)$ where ${\mathbf F}(\cdot;\alpha,\beta) = {\mathbf U}_{\tilde x} \bigodot {\mathbf c}(\adjS_d;\alpha) {\mathbf c}^T_x(\adjS_x;\beta)$ and $\bigodot$ denote element-wise product operation. Recall that$\eigenU_{\tilde x} = \eigenU^T_d {\mathbf Q}$ where ${\mathbf Q}$ is the eigen matrix of ${\mathbf X}$. ${\mathbf c}(\cdot)$ and ${\mathbf c}_x(\cdot)$ denote vectors of diagonal entries. Since we are interested in graph adaptation primarily in this work, we used fixed $\beta$ (i.e., fixed low dimensional node features). The reduced dimensionality of node features was set to $\min(m,2048)$ where $m$ is the dimension of features in ${\mathbf X}$. We next present details regarding choices of adaptation functions. 
%This network performs neighbourhood aggregation, but, with the graph adapted based on the supervised task. %As discussed in the paper, neighborhood aggregation couples the interactions of both the graph and the features in a certain way. This can be detrimental in heterophilic datasets. In the \gprgnn~section in the main paper, we define a graph operator ${\mathbf G} = \sum_{k=0}^K \alpha_k {\mathbf A}^k$. We observe that the monomial corresponding to $k=0$, is independent of graph features (${\mathbf A}$). In similar spirit, to effectively utilize the provided node features, we tried augmenting the transformed node features to the hidden layer outputs of the MLP. We provide additional results in the Section \ref{section3} to study the effect of this component.

\par

\textbf{Adaptation Functions.} The adaptation functions can take the form $C(\adjSv;\alpha) = g_s(\alpha_1) \adjSv^{g_e(\alpha_2)}$. We experimented with two different adaptation functions.
\begin{enumerate}
    \item $C_{1}(\adjSv^{i};\alpha_{1}^{i}, \alpha_{2}) = \Big( \frac{1}{1+\exp(\alpha_{1}^{i})} \Big)*(\adjSv^i)^{\max(0,\alpha_2)}$. Each eigenvalue $\adjSv^i$, is adapted by an individual scaling coefficient $g_s(\cdot)$, but the exponentiation coefficient $g_e(\cdot)$ is shared across all the eigenvalues. $g_s(\cdot)$ lies in the range $[0,1]$. Combined with suitable choice of $g_c(\cdot)$, different parts of the eigen spectrum can be suppressed or enhanced, as needed for the supervised task.
    \item $C_{2}(\adjSv^{i};\alpha_{1}^{i}, \alpha_{2}^{i}) = \max(0,\alpha_{1}^{i}) *(\adjSv^i)^{\max(0,\alpha_{2}^{i})}$. Each eigenvalue $\adjSv^{i}$, is adapted by individual scaling and exponentiation coefficients $g_s(\cdot)$ and $g_e(\cdot)$ respectively. $g_s(\cdot)$ is in the range $[0,\infty)$. This function also has the ability to suppress and enhance different parts of the eigen spectrum. %This means that each eigenvalue can be suppressed (when $g_s(\cdot)$ < 1) and amplified (when $g_s(\cdot)$ > 1). Amplification is particularly desirable when the magnitude of the eigenvalues are low. 
    Unlike the above adaptation function, we learn individual exponentiation parameters $g_e(\cdot)$ for each eigenvalue which makes this adaptation function very powerful.
\end{enumerate}

Note that even though that the adaptation function $C_{2}$ is powerful as it learns an exponentiation coefficient for each eigenvalue, it may be overparameterized for some datasets. This can result in overfitting, especially in limited labelled settings. Whereas, in the adaptation function $C_{1}$, we learn a global exponentiation coefficient and this may be insufficient for some datasets. Therefore, the above two functions operate in two extremes. To get the best of both these parameterizations, we propose \regeigeneigenet.

\par 
\textbf{\regeigeneigenet}. To reduce the number of learnable parameters, \regeigeneigenet~partitions the eigenvalues into several contiguous bins and uses shared parameters for each of the bins. This partitioning is done for both $\alpha_1^i$ and $\alpha_2^i$ discussed in the aforementioned adaptation functions. We treat the number of bins as a hyper-parameter and sweep it in the range [10\% of the no. of. nodes, 90\% of the no. of nodes]. For \regeigeneigenet~models, there is an additional $l_1$ weight regularization applied to $g_s$ and swept in the range [1e-3, 1e3] in logarithmic steps.   

%\textbf{Eigen-Concat Networks}. As described in the main paper, these models decouple the graph and the given features, leading to effective utilization of both these components in heterophilic datasets. We optimize $f(\eigenU C(\adjS;\alpha)||X)$, where $X$ is the original node features. 

In our experiments, we feed embedding outputs of all our networks to a fully-connected non-linear network with a single hidden layer with \textsc{ReLu} activation and \textsc{Softmax} final layer. We observed that using a scaled $\eigenU_d$ helps to get improved performance for datasets like Chameleon and Squirrel (with a scaling value, $s=10$). We treated the two adaptation functions described above as hyperparameters.

All models use the Adam optimizer~\cite{adam}. For our proposed models that involve learning, we set early stopping to 30 and maximum number of epochs to 300. We utilize learning rate with decay, with decay factor set to 0.99 and decay frequency set to 50. All our experiments were performed on a machine with Intel Xeon 2.60Ghz processor, 112GB Ram, Nvidia Tesla P-100 GPU with 16GB of memory, python 3.6, and Tensorflow 1.15\citep{tensorflow}. We used Optuna \citep{optuna} to optimize the hyperparameter search.

\subsection{Experiments on Heterophilic Datasets}
\label{hetero_exps}

\renewcommand{\arraystretch}{1.2}

\begin{table}[ht]
\centering
\resizebox{\textwidth}{!}{

\begin{tabular}{ccccccccc}
\hline
\textbf{Dataset} &
  \textbf{Texas} &
  \textbf{Wisconsin} &
  \textbf{Actor} &
  \textbf{Squirrel} &
  \textbf{Chameleon} &
  \textbf{Crocodile} &
  \textbf{Cornell} \\ \hline
\multicolumn{1}{c|}{\textbf{Homophily level}} & 0.11 & 0.21 & 0.22  & 0.22   & 0.23  & 0.26   & 0.30 \\
\multicolumn{1}{c|}{\textbf{\#Nodes}}         & 183  & 251  & 7600  & 5201   & 2277  & 11631  & 183  \\
\multicolumn{1}{c|}{\textbf{\#Edges}}         & 492  & 750  & 37256 & 222134 & 38328 & 191506 & 478  \\
\multicolumn{1}{c|}{\textbf{\#Features}}      & 1703 & 1703 & 932   & 2089   & 500   & 500    & 1703  \\
\multicolumn{1}{c|}{\textbf{\#Classes}}       & 5    & 5    & 5     & 5      & 5     & 6      & 5     \\
\multicolumn{1}{c|}{\textbf{\#Train/Val/Test}} &
  87/59/37 &
  120/80/51 &
  3648/2432/1520 &
  2496/1664/1041 &
  1092/729/456 &
  120/180/11331 &
  87/59/37  \\ \hline
\end{tabular}

}
\vspace{1mm}
\caption{Datasets Statistics}
\label{tab:stats_table}
\end{table}
% \begin{enumerate}
%     \item Main table - comparison with baselines on all datasets. Perhaps it may help to organize/group tables as we had done in the OneNote instead of giving a single long table. 
%     \item Make key observations: (a) using SVD(A) only and X only based models, (b) concat models and (c) Aggregation models. 
%     \item Make observations related to with and without tuning. 
%     \item Make observations with other competitive baselines. 
%     \item Make observations related to concat and aggregation models.
%     \item Make observations pertaining to heterophily and homophily datasets.
%     \item In each of the above items, try to give some reasoning behind them. 
% \end{enumerate}

\textbf{Datasets.} We evaluate on seven heterophilic datasets to show the effectiveness of our approach. Detailed statistics of the datasets used are provided in Table~\ref{tab:stats_table}. We borrowed \textbf{Texas}, \textbf{Cornell}, \textbf{Wisconsin} from WebKB\footnote{http://www.cs.cmu.edu/afs/cs.cmu.edu/project/theo-11/www/wwkb}, where nodes represent web pages and edges denote hyperlinks between them. \textbf{Actor} is a co-occurence network borrowed from~\cite{actordataset}, where nodes correspond to an actor, and and edge represents the co-occurrence on the same Wikipedia page. \textbf{Chameleon}, \textbf{Squirrel}, and \textbf{Crocodile} are borrowed from~\cite{chameleondataset}. Nodes correspond to web pages and edges capture mutual links between pages. For all benchmark datasets, we use feature vectors, class labels from~\cite{supergat}. For datasets in (Texas, Wisconsin, Cornell, Chameleon, Squirrel, Actor), we use 10 random splits (48\%/32\%/20\% of nodes for train/validation/test set) from~\cite{geomgcn}. For Crocodile, we create 10 random splits following~\cite{supergat}.

\begin{table}[ht]
\resizebox{\textwidth}{!}{%
\begin{tabular}{@{}cccccccc@{}}
\toprule
                            & \textbf{Texas} & \textbf{Wisconsin} & \textbf{Actor} & \textbf{Squirrel} & \textbf{Chameleon} & \textbf{Crocodile} & \textbf{Cornell} \\ \midrule
\textbf{LR}       & 81.35 (6.33) & 84.12 (4.25) & 34.70 (0.89) & 34.73 (1.39) & 48.25 (2.67) & 48.25 (2.67) & 83.24 (5.64) \\
\textbf{MLP}      & 81.24 (6.35) & 84.43 (5.36) & \textbf{36.06 (1.11)} & 35.38 (1.38) & 51.64 (1.89) & 54.47 (1.99) & 83.78 (5.80) \\
\textbf{\sgcn}     & 62.43 (4.43) & 55.69 (3.53) & 30.44 (0.91) & 45.72 (1.55) & 60.77 (2.11) & 51.54 (1.47) & 62.43 (4.90) \\
\textbf{\gcn}      & 61.62 (6.14) & 53.53 (4.73) & 30.32 (1.05) & 46.04 (1.61) & 61.43 (2.70) & 52.34 (2.61) & 62.97 (5.41) \\
\textbf{\supergat} & 61.08 (4.97) & 56.47 (3.90) & 29.32 (1.00) & 31.84 (1.26) & 43.22 (1.71) & 52.41 (1.92) & 57.30 (8.53) \\
\textbf{\geomgcn} & 67.57*       & 64.12*       & 31.63*       & 38.14*       & 60.90*       & NA           & 60.81*       \\
\textbf{\hhgcn}              & \textbf{84.86 (6.77)}*  & 86.67 (4.69)*      & 35.86 (1.03)*  & 37.90 (2.02)*     & 58.40 (2.77)       & 53.17 (1.21)       & 82.16 (4.80)*    \\
\textbf{\fagcn}    & 82.43 (6.89) & 82.94 (7.95) & 34.87 (1.25) & 42.59 (0.79) & 55.22 (3.19) & 54.35 (1.05) & 79.19 (9.79) \\
\textbf{\appnp}    & 81.89 (5.85) & 85.49 (4.45) & 35.93 (1.04) & 39.15 (1.88) & 47.79 (2.35) & 53.13 (1.93) & 81.89 (6.25) \\
\textbf{\gprgnn}  & 81.35 (5.32) & 82.55 (6.23) & 35.16 (0.90) & 46.31 (2.46) & 62.59 (2.04) & 52.71 (1.84) & 78.11 (6.55) \\ \midrule
\multicolumn{8}{c}{\textbf{Eigen Network Models}}                                                                          \\ \midrule
\textbf{\eigennet}    & 58.92 (3.78) & 53.14 (4.84) & 25.37 (0.88) & 54.62 (1.5)  & \textbf{67.28 (2.21)} & 45.54 (2.08) & 57.30 (5.10) \\
\textbf{\eigeneigennet}      & 82.70 (6.42)   & 82.75 (4.79)       & 35.04 (0.91)   & 57.11 (1.94)      & 65.79 (1.16)       & 54.51 (1.93)       & 77.30 (6.19)     \\
\textbf{\regeigeneigenet} & 84.05 (5.76)   & \textbf{89.80 (4.22)}       & 34.84 (0.53)   & \textbf{57.61 (1.92)}      & 66.45 (2.77)       & \textbf{55.03 (2.12)}       & \textbf{84.86 (4.80)} \\   
\textbf{\eigenconcat}       & 78.11 (3.72)   & 85.69 (4.81)       & 34.75 (0.83)   & 53.66 (1.76)      & 66.51 (1.21)       & 54.20 (1.61)       & 80.81 (6.10)     \\ \bottomrule
\end{tabular}%
}
\caption{Comparison of various Eigen Networks with Baselines. The results marked with "*" are obtained from the corresponding paper. }
\label{tab:main_table}
\end{table}

We propose the following models i) \eigennet: We observe from Table~\ref{tab:main_table} that \eigennet~models, which only rely on topological information, perform better than baselines on Squirrel and Chameleon datasets. This indicates that graph features by themselves are useful for a few datasets. \par
ii) \eigeneigennet: these models extend \eigennet~models to incorporate aggregation. In our experiments, we restrict adaptation to only the topology by setting $\beta$ to 1. This also allows us to study the effect of graph adaptation. \eigeneigennet~models are powerful than \eigennet~models mainly because they are also able to churn out information from the feature space. In comparison against baselines, this approach clearly outperforms baseline models. The greatest gains can be noted in Squirrel and Chameleon datasets with accuracy gains of up to 11\%. We believe that graph adaptation function is the reason for performance gains as it is able to highlight important signals in the topology that can complement the aggregation. We can empirically observe that this proposed way of aggregation is effective for heterophilic datasets. 
\par
iii) \regeigeneigenet~is a regularized version of \eigeneigennet, which is able to gain further improvements by reducing the number of learning parameters. In specific, we observe that \regeigeneigenet~model consistently outperform across several datasets. We empirically observe that grouping contiguous eigenvalues and learning shared coefficients provides a regularizing effect and improves model's generalizability. It can be inferred from Table~\ref{tab:main_table} that although several baselines were proposed to address heterophily in graphs, there is no single baseline that consistently achieves good performance across the benchmark datasets. \par
iv) \eigenconcat: these models deviate from the popular aggregation scheme and offers an effective solution that leverage signals from the given topology and features. For instance, on Wisconsin and Chameleon, these models outperform non-graph based methods and several aggregation methods including \gcn, \sgcn, \supergat~and even \geomgcn. With respect to our proposed models, we make a global observation that graph adaptation persistently improves performance with gains of up to 11\% as showcased in Table~\ref{tab:adaptation}.

% We observe in Table~\ref{tab:main_table} that our proposed models outperform baseline models on several benchmark datasets with accuracy gains of up to 20\%. On the rest datasets our models show competitive performance. It can be inferred from Table~\ref{tab:main_table} that simple models like Logistic Regression (LR) and MLP outperform baseline models like \supergat and \geomgcn, which were designed for heterophilic settings, on several datasets. It can be noted from Table~\ref{tab:main_table} that our proposed aggregation based methods also offer competitive performance against baseline models and also outperform them with accuracy gains up to 3\% on several datasets.

% \textbf{Concat v/s Aggregation for Heterophilic Settings}
% We observe that our concatenation models, which concatenate graph features and given node features, perform better than aggregation models. 

\subsection{Experiments on Homophilic Datasets}
\label{homo_exps}

Our paper mainly focuses on heterophilic datasets. However, it is also important to understand the performance of our model on homophilic datasets. Towards that end, we ran experiments on some homophilic datasets. The datasets used and their corresponding statistics are available in Table~\ref{tab:StatsTable}. We borrowed \textbf{Cora}, \textbf{Citeseer}, and \textbf{Pubmed} datasets and the corresponding train/val/test set splits from \cite{geomgcn}. The remaining datasets were borrowed from \cite{supergat}. We follow the same dataset setup mentioned in \cite{supergat} to create 10 random splits for each of these datasets. 
% We also experiment with a slightly larger heterophilic dataset, \textbf{Flickr}. We use the publicly available fixed split for this dataset.

\begin{table}[H]
\resizebox{\textwidth}{!}{%
\begin{tabular}{@{}c|cccccccc@{}}
\toprule
\textbf{Statistics} & \textbf{Flickr} & \textbf{Cora-Full} & \textbf{Wiki-CS} & \textbf{Citeseer} & \textbf{Pubmed} & \textbf{Cora} & \textbf{Computer} & \textbf{Photos} \\ \midrule
\textbf{Homophily Score}    & 0.32   & 0.59  & 0.68   & 0.74  & 0.80   & 0.81  & 0.81   & 0.85   \\
\textbf{Number of Nodes}    & 89250  & 19793 & 11701  & 3327  & 19717  & 2708  & 13752  & 7650   \\
\textbf{Number of Edges}    & 989006 & 83214 & 302220 & 12431 & 108365 & 13264 & 259613 & 126731 \\
\textbf{Number of Features} & 500    & 500   & 300    & 3703  & 500    & 1433  & 767    & 745    \\
\textbf{Classes}            & 7      & 70    & 10     & 6     & 3      & 7     & 10     & 8      \\
\textbf{\#Train}            & 44625  & 1395  & 580    & 1596  & 9463   & 1192  & 200    & 160    \\
\textbf{\#Validation}       & 22312  & 2049  & 1769   & 1065  & 6310   & 796   & 300    & 240    \\
\textbf{\#Test}             & 22313  & 16349 & 5847   & 666   & 3944   & 497   & 13252  & 7250   \\ \bottomrule
\end{tabular}%
}
\caption{Datasets Statistics}
\label{tab:StatsTable}
\end{table}

\textbf{Performance on Homophilic v/s Heterophilic Datasets.} The performance results of the various baselines and our approach is given in Table~\ref{tab:HomophilyTable}. We observe that on homophilic datasets, \eigeneigennet~which is an aggregation based model, tends to perform better than \eigenconcat~models. This trend is expected; as the homophily levels of the graph increases the discord between the node features ($\textbf{X}$) and topology ($\textbf{A}$) reduces, leading to improvements in aggregation methods. Baseline models including \gprgnn, \supergat, \fagcn, and \appnp, which are also aggregation based methods, perform better on homophilic datasets. However, our proposed models are not far off.

\begin{table}[H]
\resizebox{\textwidth}{!}{%
\begin{tabular}{@{}cccccccc@{}}
\toprule
\multicolumn{1}{c|}{\textbf{Test Acc}} &
  \textbf{Cora-Full} &
  \textbf{Wiki-CS} &
  \textbf{Citeseer} &
  \textbf{Pubmed} &
  \textbf{Cora} &
  \textbf{Computer} &
  \textbf{Photos} \\ \midrule
\multicolumn{1}{c|}{\textbf{LR}}       & 39.10   (0.43) & 72.28   (0.59) & 72.22   (1.54)  & 87.00   (0.40)  & 73.94   (2.47)  & 64.92   (2.59) & 77.57   (2.29) \\
\multicolumn{1}{c|}{\textbf{MLP}}      & 43.03   (0.82) & 73.74   (0.71) & 73.83   (1.73)  & 87.77   (0.27)  & 77.06   (2.16)  & 64.95   (3.57) & 76.96   (2.46) \\
\multicolumn{1}{c|}{\textbf{GCN}}      & 45.44   (1.01) & 77.64   (0.49) & 76.47   (1.33)  & 87.86   (0.47)  & 87.28   (1.34)  & 78.16   (1.85) & 86.38   (1.71) \\
\multicolumn{1}{c|}{\textbf{SGCN}}     & 61.31   (0.78) & 78.30   (0.75) & 76.77   (1.52)  & 88.48   (0.45)  & 86.96   (0.78)  & 80.65   (2.78) & 89.99   (0.69) \\
\multicolumn{1}{c|}{\textbf{SuperGAT}} & 57.75   (0.97) & 77.92   (0.82) & 76.58   (1.59)  & 87.19   (0.50)  & 86.75   (1.24)  & \textbf{83.04   (1.02)} & 90.31   (1.22) \\
\multicolumn{1}{c|}{\textbf{GeomGCN}}  & NA             & NA             & 77.99*          & 90.05*          & 85.27*          & NA             & NA             \\
\multicolumn{1}{c|}{\textbf{H2GCN}}    & 57.83   (1.47) & OOM            & 77.07   (1.64)* & \textbf{89.59   (0.33)*} & 87.81   (1.35)* & OOM            & 91.17   (0.89) \\
\multicolumn{1}{c|}{\textbf{FAGCN}}    & 60.07   (1.43) & 79.23   (0.66) & 76.80   (1.63)  & 89.04   (0.50)  & \textbf{88.21   (1.37)}  & 82.16   (1.48) & 90.91   (1.11) \\
\multicolumn{1}{c|}{\textbf{GPR-GNN}}  & \textbf{61.37   (0.96)} & \textbf{79.68   (0.50)} & 76.84   (1.69)  & 89.08   (0.39)  & 87.77   (1.31)  & 82.38   (1.60) & 91.43   (0.89) \\
\multicolumn{1}{c|}{\textbf{APPNP}}    & 60.83   (0.55) & 79.13   (0.50) & 76.86   (1.51)  & 89.57   (0.53)  & 88.13 (1.53)    & 82.03   (2.04) & \textbf{91.68   (0.62)} \\ \midrule
\multicolumn{8}{c}{\textbf{Our Models}}                                                                                                                                  \\ \midrule
\multicolumn{1}{c|}{\textbf{EigenNetwork}} &
  43.93   (1.19) &
  61.75   (1.31) &
  65.53   (3.49) &
  81.07   (0.41) &
  80.12   (1.54) &
  64.17   (6.19) &
  74.79   (3.44) \\
\multicolumn{1}{c|}{\textbf{Eigen-EigenNetwork}} &
  56.10   (1.03) &
  77.96   (0.53) &
  76.67   (1.83) &
  \underline{\textit{89.30   (0.42)}} &
  87.10   (1.10) &
  78.86   (1.86) &
  88.50   (0.92) \\
\multicolumn{1}{c|}{\textbf{Eigen-ConcatNetwork}} &
  47.47   (0.92) &
  74.13   (0.87) &
  74.86   (1.90) &
  88.38   (0.15) &
  84.43   (1.77) &
  66.20   (3.33) &
  76.97   (2.63) \\
\multicolumn{1}{c|}{\textbf{RegEigen-EigenNetwork}} &
  \underline{\textit{58.19   (0.62)}} &
  \underline{\textit{78.31   (0.69)}} &
  \underline{\textbf{\textit{77.20   (1.36)}}} &
  89.22   (0.43) &
  \underline{\textit{87.17   (1.18)}} &
  \underline{\textit{81.06   (1.80)}} &
  \underline{\textit{89.01   (1.05)}} \\ \bottomrule
\end{tabular}%
}
\vspace{2mm}
\caption{Homophily Datasets Results. We bold the the results of the best performing models for each dataset. We underline and italicize the best performing Eigen models for ease of comparison. '*' indicates that the results were borrowed from the corresponding papers.}
\label{tab:HomophilyTable}
\end{table}

\subsection{Experiments on Large Dataset}

We additionally performed one large scale dataset experiment on \textbf{Flickr} dataset. We use the publicly available fixed split for this dataset. We show the results of all the models on \textbf{Flickr} in Table~\ref{tab:res_flickr}. We first note that \supergat~gives the best performance among all baselines followed by \gprgnn. However, three of our models: \eigennet, \eigeneigennet~and \regeigeneigenet~outperform all the baselines. Amongst our models, \eigennet~gives the best performance with 54.4\% test accuracy. Another thing to note here is that \eigenconcat~model is not far behind the other \eigennet~models and do better than \appnp~and \sgcn. This suggests that concatenation based models can offer effective alternative for aggregation based approaches.

\begin{table}[H]
\centering
\resizebox{0.75\textwidth}{!}{%
\begin{tabular}{@{}cc|cc|cc@{}}
\toprule
\textbf{Model} & \textbf{Test Acc} & \textbf{Model} & \textbf{Test Acc} & \textbf{Model} & \textbf{Test Acc} \\ \midrule
\textbf{LR}   & 46.51 & \textbf{SuperGAT} & 53.47 & \textbf{EigenNetwork}          & \textbf{54.4}  \\
\textbf{MLP}  & 46.93 & \textbf{GeomGCN}  & NA    & \textbf{Eigen-EigenNetwork}    & 53.78 \\
\textbf{GCN}  & 53.4  & \textbf{H2GCN}    & OOM   & \textbf{Eigen-ConcatNetwork}   & 51.78 \\
\textbf{SGCN} & 50.75 & \textbf{FAGCN}    & OOM   & \textbf{RegEigen-EigenNetwork} & 53.83 \\
              &       & \textbf{GPR-GNN}  & 52.74 &                                &       \\
              &       & \textbf{APPNP}    & 50.33 &                                &       \\ \bottomrule
\end{tabular}%
}
\vspace{2mm}
\caption{Results on Flickr Dataset. We report Test Accuracy on the standard split for all the models.}
\label{tab:res_flickr}
\end{table}

\subsection{Analysis}
\label{analysis}
\begin{figure}[H]
    \centering
    \subfigure[Crocodile]{
        \label{fig: crocodile_eig}
        \includegraphics[width=0.3\textwidth]{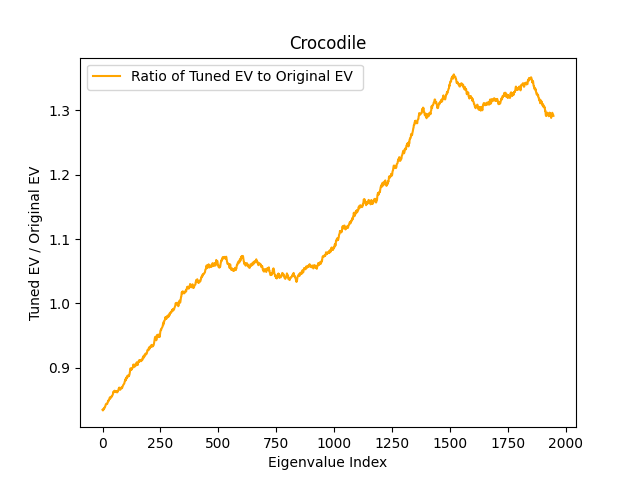}
    }
    \subfigure[Chameleon]{
        \label{fig:chameleon_eig}
        \includegraphics[width=0.3\textwidth]{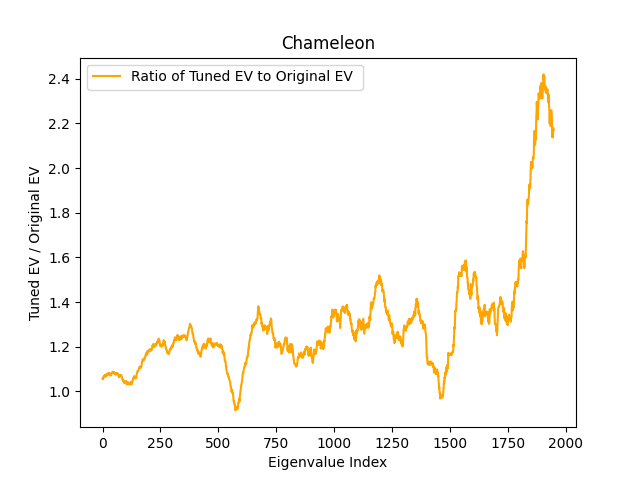}
    }
    \subfigure[Squirrel]{
        \label{fig:squirrel_eig}
        \includegraphics[width=0.3\textwidth]{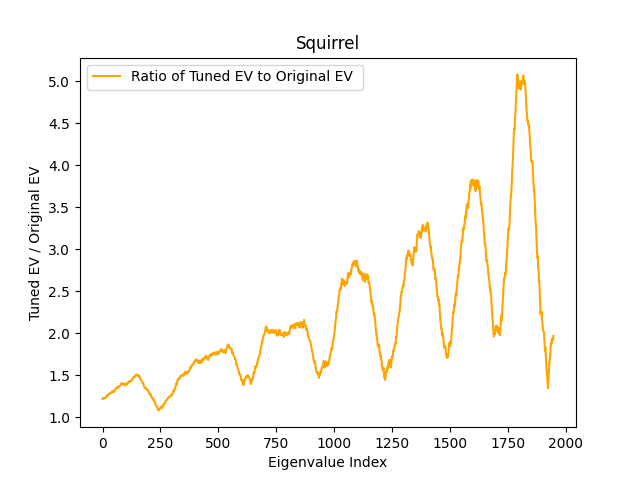}
    }
    \caption{We plot the ratio of the tuned eigenvalues to the original eigenvalues of the graph.}
    \label{fig:eig_ratio}
\end{figure}

\textbf{Effect of Adaptation:} We carry out an ablation study in Table \ref{tab:adaptation} by freezing the original eigenvalues. We observe that adaptation helps in most of the datasets for all the proposed \eigennet~variants. There is a significant improvement for \eigeneigennet~over the freezed variant in datasets like Texas, Wisconsin and Cornell. As observed in Section \ref{sec:eigenvectoranalysis}, there exists different weightings for the eigenvalues which are more useful for the supervised task. Additionally in Figure~\ref{fig:eig_ratio}, we plotted the ratio of the adapted eigenvalues to that of the original eigenvalues of the graph. We see that different regions of eigenvalues are given high weights as seen in Crocodile and Squirrel plots. It might be difficult for models like \gprgnn~to learn such behaviour with a polynomial. Our proposed approaches are able to learn such behavior, which is also reflected in the performance where we gain 3\% over \gprgnn~on Crocodile and up to 11\% on Squirrel.

\begin{table}[ht]
\resizebox{\textwidth}{!}{%
\begin{tabular}{@{}c|ccccccc@{}}
\toprule
\textbf{} & \textbf{Texas} & \textbf{Wisconsin} & \textbf{Actor} & \textbf{Squirrel} & \textbf{Chameleon} & \textbf{Crocodile} & \textbf{Cornell} \\ \midrule
\textbf{\eigennet~ w/o Adaptation}         & 56.76 (4.83) & 50.78 (4.92) & 25.24 (0.84) & 53.67 (1.4)  & 65.61 (1.63) & 44.96 (1.78) & 55.14 (7.47) \\
\textbf{\eigennet}     & \textbf{58.92 (3.78)} & \textbf{53.14 (4.84)} & \textbf{25.37 (0.88)} & \textbf{54.62 (1.5)}  & \textbf{67.28 (2.21)} & \textbf{45.54 (2.08)} & \textbf{57.30 (5.10)} \\ \midrule
\textbf{\eigenconcat~ w/o adaptation} & \textbf{78.38 (5.92)} & 84.9 (5.12)  & \textbf{34.93 (0.63)} & 47.57 (1.75) & 63.79 (2.14) & 54.42 (1.62) & \textbf{81.89 (5.41)} \\
\textbf{\eigenconcat}  & 78.11 (3.72) & \textbf{85.69 (4.81)} & 34.75 (0.83) & \textbf{53.66 (1.76)} & \textbf{66.51 (1.21)} & \textbf{54.20 (1.61)} & 80.81 (6.10) \\ \midrule
\textbf{\eigeneigennet~ w/o adaptation}  & 65.41 (6.71) & 69.41 (4.04) & 27.66 (0.78) & 56.01 (1.48) & 60.29 (2.54) & 53.89 (0.96) & 67.03 (4.49) \\
\textbf{\eigeneigennet} & \textbf{82.70 (6.42)} & \textbf{82.75 (4.79)} & \textbf{35.04 (0.91)} & \textbf{57.11 (1.94)} & \textbf{65.79 (1.16)} & \textbf{54.51 (1.93)} & \textbf{77.30 (6.19)} \\ \bottomrule
\end{tabular}%
}

\caption{Effect of Eigen Value Adaptation}
\label{tab:adaptation}
\end{table}

\textbf{Node and Graph Features:} We conduct a study to answer the question: \textit{Are both node and graph features helpful and needed?} We experimented using three different models, fully connected multi-layer nonlinear network (MLP) with a single hidden layer with hidden dimensions swept in the range [16,32,64], \eigennet~and \eigenconcat. MLP uses only the node features, \eigennet~ uses only the graph/topological information. \eigenconcat~ leverages both topological and node features. In Table \ref{tab:node_graph}, we observe across several datasets that \eigenconcat~ model performs better than either of the individual models. We see ~4\% improvement in \textbf{Cora-Full} and \textbf{Cora} datasets over the best performing individual model (\eigennet). This highlights the fact that both these sources are helpful and useful to build simple and efficient \eigenconcat~ models.

\begin{table}[H]
\centering
\resizebox{\textwidth}{!}{%
\begin{tabular}{@{}c|ccccccc@{}}
\toprule
\textbf{Test Acc}     & \textbf{Cora-Full} & \textbf{Wiki-CS} & \textbf{Citeseer} & \textbf{Pubmed} & \textbf{Cora} & \textbf{Computer} & \textbf{Photos} \\ \midrule
\textbf{MLP}          & 43.03 (0.82)       & 73.74 (0.71)     & 73.83 (1.73)      & 87.77 (0.27)    & 77.06 (2.16)  & 64.95 (3.57)      & 76.96 (2.46)    \\
\textbf{EigenNetwork} & 43.93 (1.19)       & 61.75 (1.31)     & 65.53 (3.49)      & 81.07 (0.41)    & 80.12 (1.54)  & 64.17 (6.19)      & 74.79 (3.44)    \\
\textbf{Eigen-ConcatNetwork} &
  \textbf{47.47 (0.92)} &
  \textbf{74.13 (0.87)} &
  \textbf{74.86 (1.90)} &
  \textbf{88.38 (0.15)} &
  \textbf{84.43 (1.77)} &
  \textbf{66.20 (3.33)} &
  \textbf{76.97 (2.63)} \\ \bottomrule
\end{tabular}%
}
\vspace{2mm}
\caption{Complementary Information: Node and Graph Features}
\label{tab:node_graph}
\end{table}

\textbf{Using Node Features with Aggregation Models:} Recall that \gprgnn~ propagation model uses the graph operator, ${\mathbf G} = \sum_{k=0}^K \alpha_k {\mathbf A}^k$. We observe that the monomial corresponding to $k=0$, is independent of neighborhood aggregation using powers of (${\mathbf A}$). This could be a very useful signal to include, in particular, when the node features are of high quality and possess relatively strong discriminative power. \eigeneigennet~ models do not consume the node features directly, but, can be easily included. In Table \ref{tab:order0}, we report performance numbers on a few datasets, with and without explicitly augmenting Order-0 monomial to \eigeneigennet. We see ~7.5\% and ~5\% improvement in \textbf{Actor} and \textbf{Pubmed}. %In these datasets MLP on node features performs reasonably well relative to other models, this augmentation helps. We observe gains in few other datasets too. 
However, this improvement was not observed on several other datasets.

\begin{table}[H]
\centering
\resizebox{0.45\textwidth}{!}{%
\begin{tabular}{@{}c|cc@{}}
\toprule
\textbf{Eigen-EigenNetwork}       & \textbf{Actor} & \textbf{Pubmed} \\ \midrule
\textbf{Without Order-0 Monomial} & 27.51 (0.91)   & 84.24 (0.50)    \\
\textbf{With Order-0 Monomial}    & \textbf{35.04 (0.91)}   & \textbf{89.30 (0.42) }   \\ \bottomrule

\end{tabular}%
}
\vspace{1mm}
\caption{Effect of Order-0 Monomial on selected datasets.}

\label{tab:order0}
\end{table}

% \textbf{Effect of varying training labels:} Table~\ref{tab:main_table} shows the effectiveness of our proposed model \regeigeneigenet. To further understand the role of weight tying, we study the effect of varying training labels for \eigeneigennet~and~\regeigeneigenet. Figure~\ref{fig:varyLabelPlots} shows the effect of weight tying on Chameleon and Squirrel datasets as we decrease the number of training labels. We further observe that even with half the number of training samples, there is only a marginal drop in performance. 

% \textbf{Performance on Homophilic v/s Heterophilic Datasets:} The performance results of the various baselines and our approach is given in Table~\ref{tab:HomophilyTable}. We observe that on homophilic datasets, \eigeneigennet~which is an aggregation based model, tends to perform better than \eigenconcat~models. This trend is expected; as the homophily levels of the graph increases the discord between the node features ($\textbf{X}$) and topology ($\textbf{A}$) reduces, leading to improvements in aggregation methods. Baseline models including \gprgnn, \supergat, \fagcn, and \appnp, which are also aggregation based methods, perform better on homophilic datasets. However, our proposed models are not far off.

\begin{figure}[H]
    \centering
    \subfigure[Chameleon]{
        \label{fig:varEigChameleon}
        \includegraphics[width=0.4\textwidth]{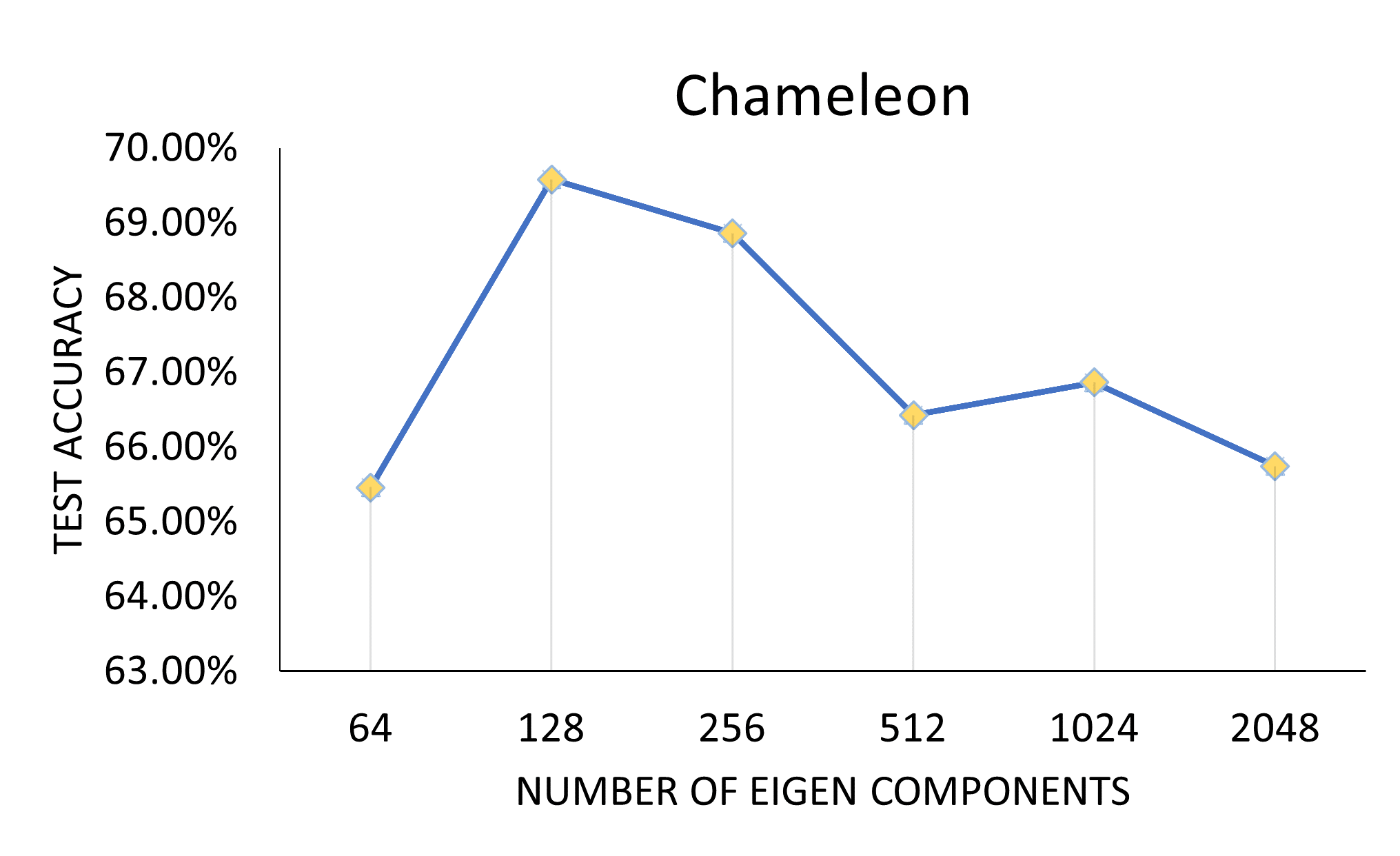}
    }
    \subfigure[Squirrel]{
        \label{fig:varEigSquirrel}
        \includegraphics[width=0.4\textwidth]{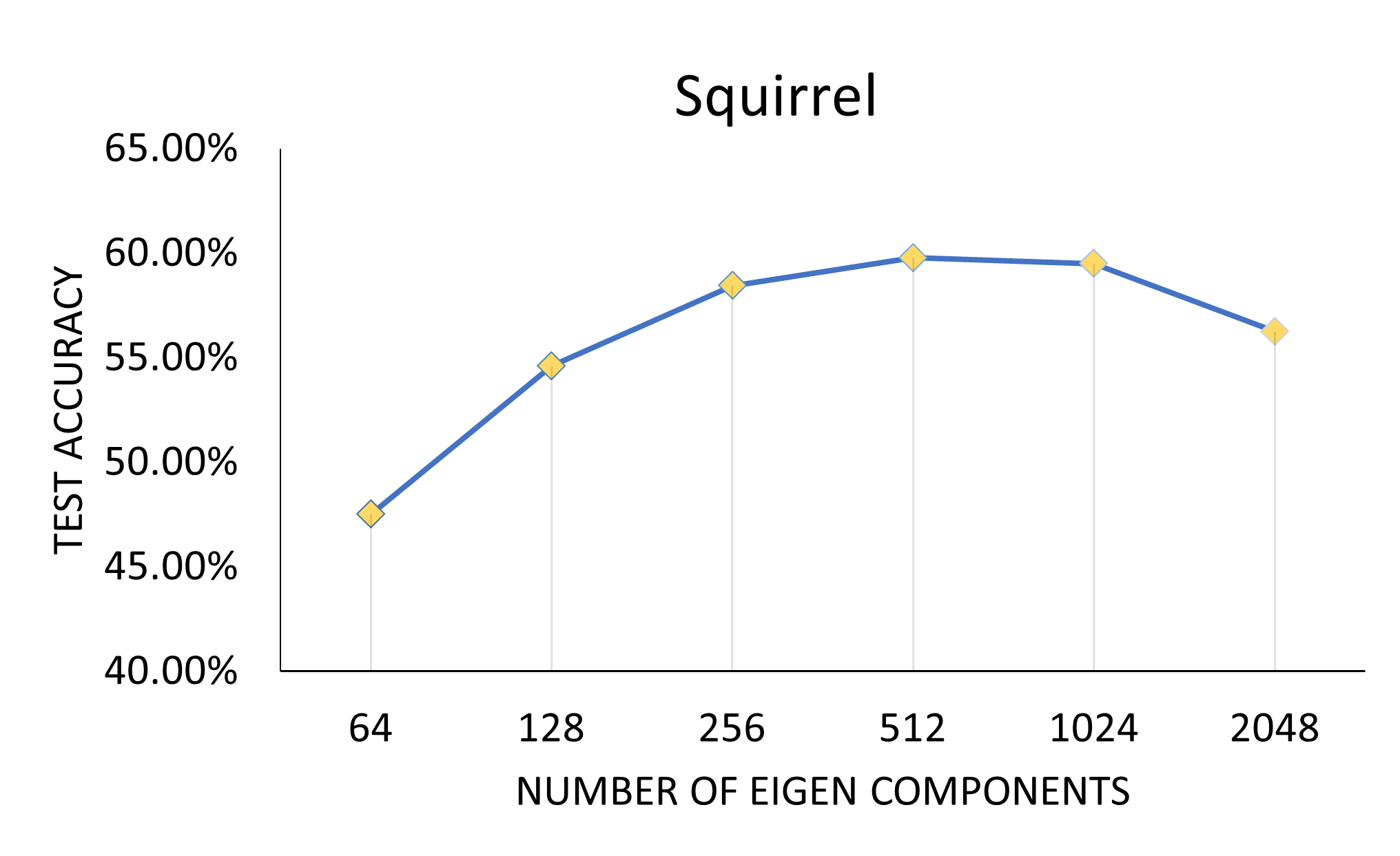}
    }
    \\
    \subfigure[Cora]{
        \label{fig:varEigCora}
        \includegraphics[width=0.4\textwidth]{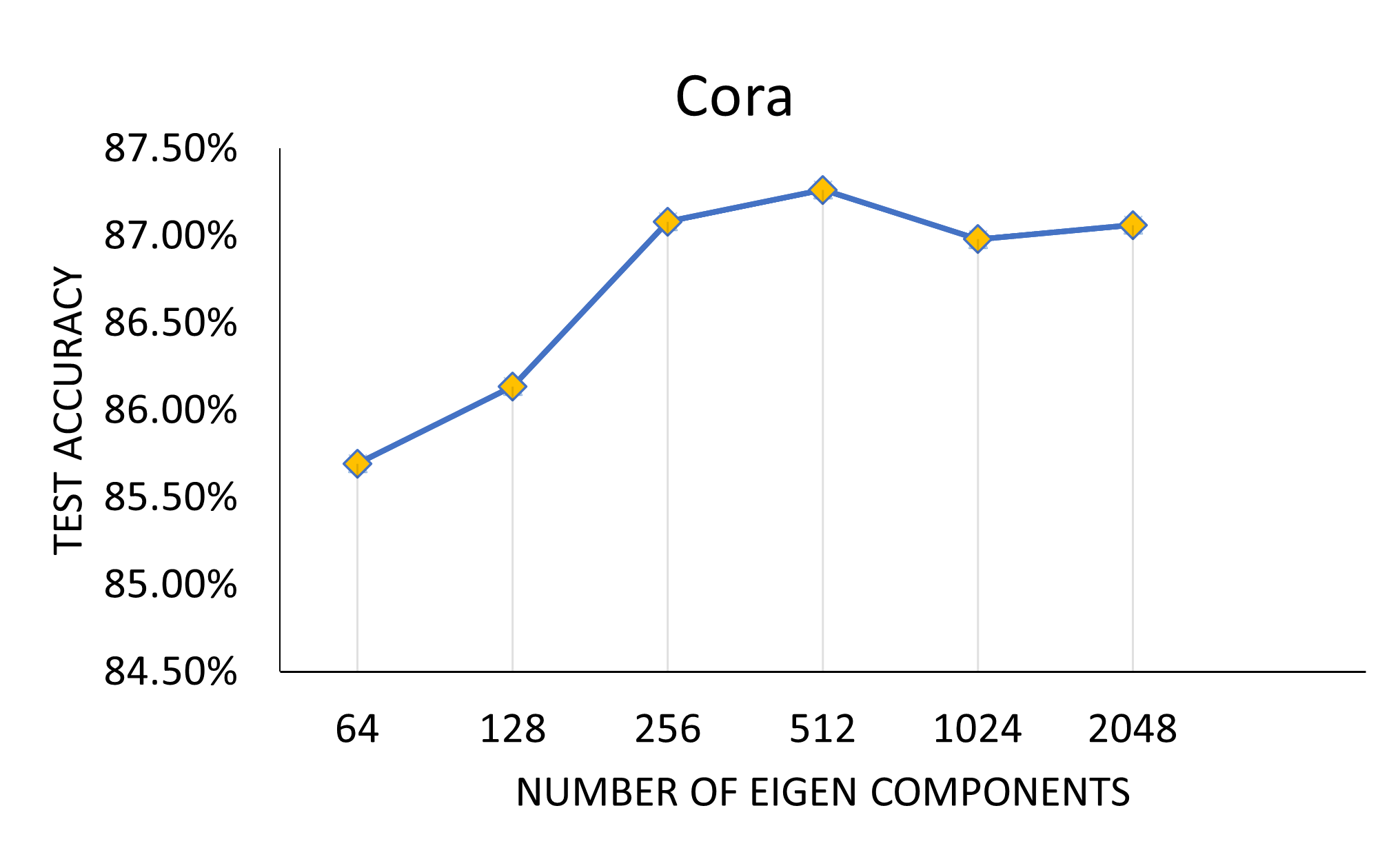}
    }
    \subfigure[Cora-Full]{
        \label{fig:varEigCoraFull}
        \includegraphics[width=0.4\textwidth]{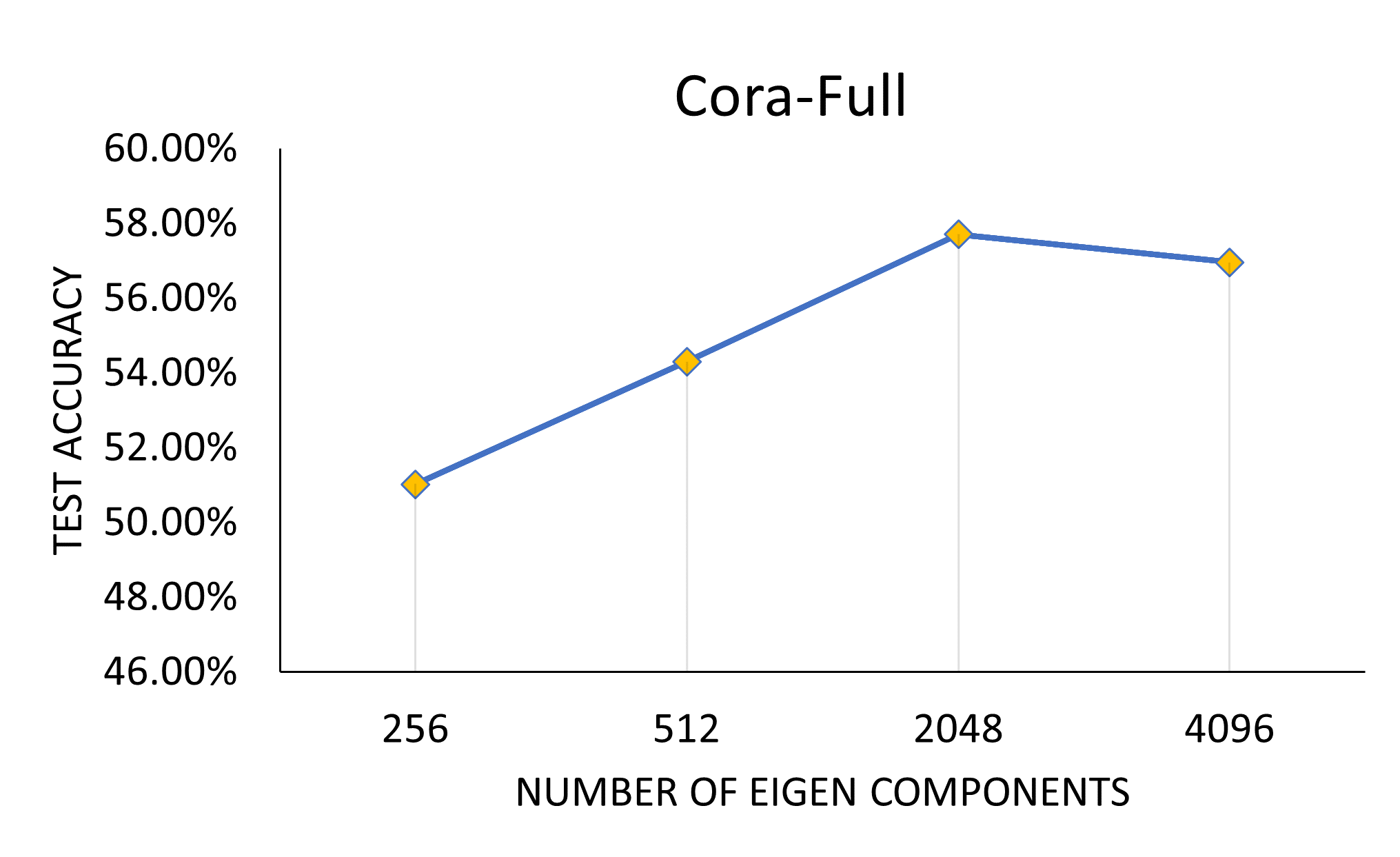}
    }
    \caption{Varying Eigen Components Plots}
    \label{fig:varyEigPlots}
\end{figure}

\textbf{Are \eigennet~models lacking on Homophilic Datasets?}. Neighborhood aggregation based methods work with $\adj$ directly. It means that they have access to the entire eigenvalue spectrum. For example, in \gcn~when we multiply $\tilde{\adj}$ with $\feat$, we get
\begin{gather}
    \tilde{\adj} \feat = \sum_{i}\adjSv_i \eigenUc_i\eigenUc_i^{T}\feat
\end{gather}
thereby leveraging the entire eigenvalue spectrum. However, in our \eigennet~models, we restrict the number of components to only top few eigen components. We believe this may be a cause for the gap in performance on homophilic datasets.
% In homophilic datasets, we hypothesize that there are useful signals throughout the spectrum . In our models, as we restrict to a fewer eigen components, we lose out on those signals. 
To study this, we do the following experiment - we vary the number of eigen components and observe the test performance. The experimental results are given in Figure~\ref{fig:varyEigPlots}. On heterophilic datasets, \textbf{Chameleon} and \textbf{Squirrel}, we do not see any observable trend in performance. However, on homophilic datasets, \textbf{Cora} and \textbf{Cora-Full}, the performance does go up with the number of components. However, it also seems to drop occasionally. For example, we observe that in \textbf{Cora-Full}, the test accuracy dropped by 2\% going from 2048 to 4096 components. This suggests that while number of components may have a role to play in the lower performance on homophilic datasets, that alone may not be the only contributor. This requires further investigation and we plan to make it a part of our future work.

\begin{figure}
    \centering
    \subfigure[Chameleon]{
        \label{fig:chameleon_vary}
        \includegraphics[width=0.4\textwidth]{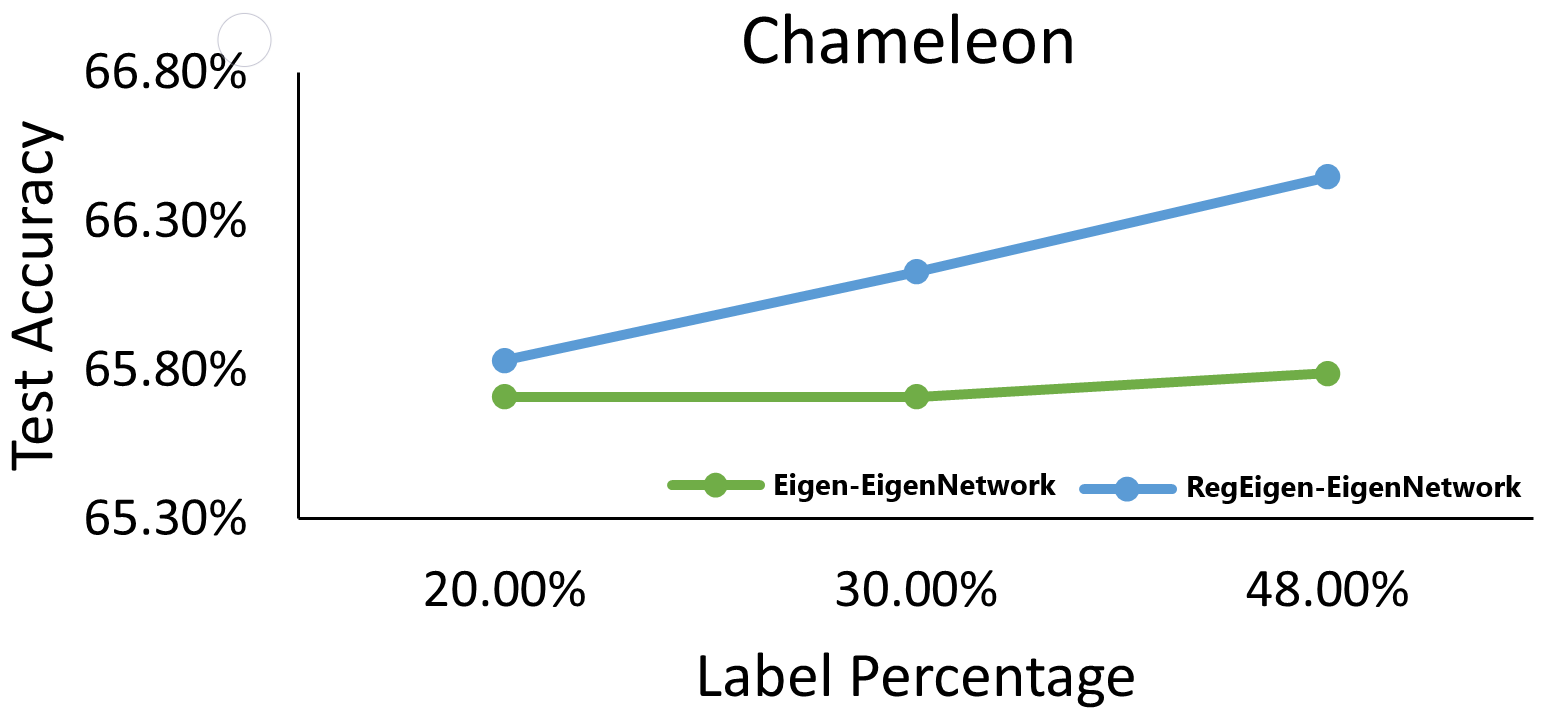}
    }
    \subfigure[Squirrel]{
        \label{fig:squirrel_vary}
        \includegraphics[width=0.4\textwidth]{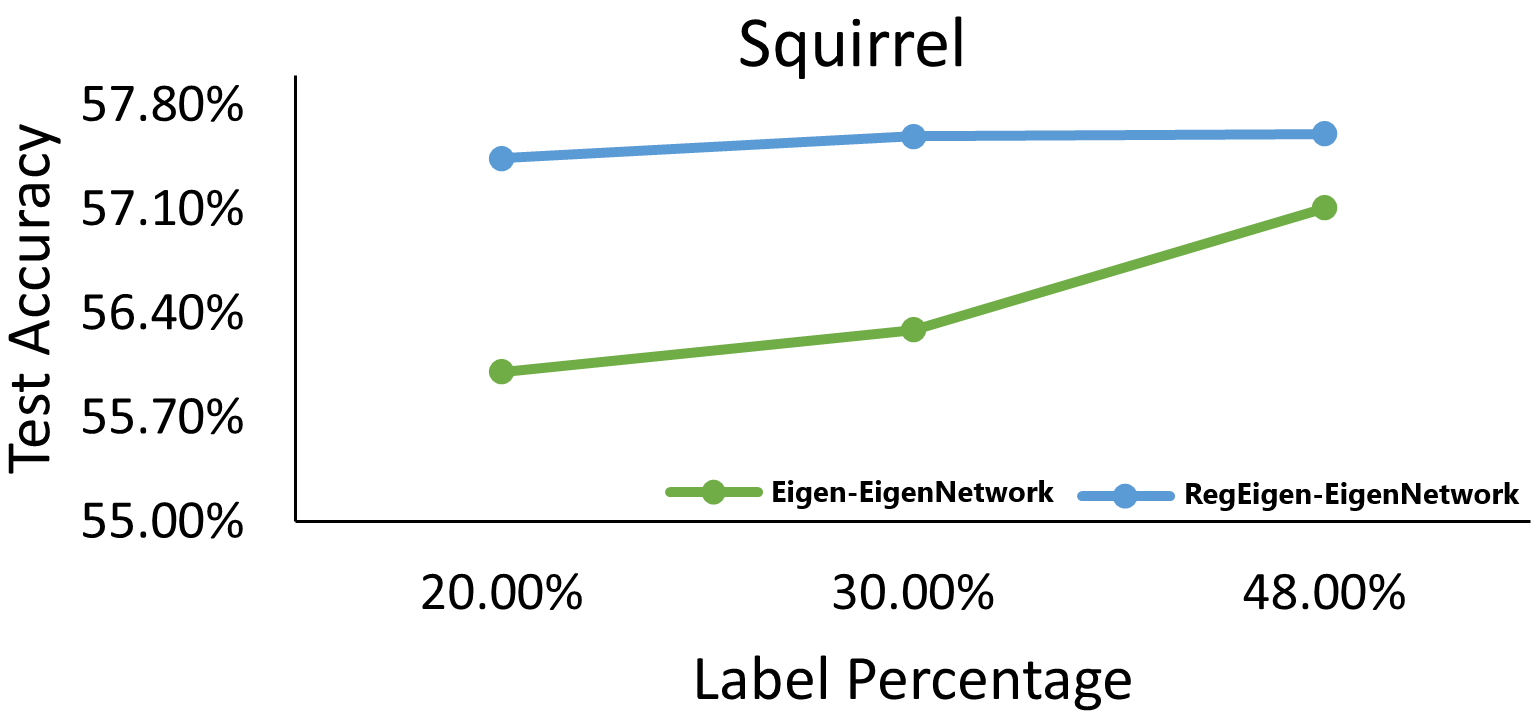}
    }
    \caption{Varying Label Plots}
    \label{fig:varyLabelPlots}
\end{figure}

\textbf{Utility of weight-tying in limited labeled data setting:} To analyze the importance of this regularization, we study the effect of varying the training set size for \eigeneigennet~and \regeigeneigenet~and plot them in Figure~\ref{fig:varyLabelPlots}. On \textbf{Squirrel}, we observe that even with 20\% labeled data, \regeigeneigenet~as good as having 48\% labeled data. On \textbf{Chameleon}, we observe something even more interesting. The base \eigeneigennet~model does not improve much in performance even with increased labeled data. However, the \regeigeneigenet~model continue to improve in performance with increased labeled data. Weight-tying is dependent on finding good partitions of eigenvalues which is currently tuned as a hyperparameter. We believe that in \textbf{Chameleon}, with increased labeled data, the model does better on validation set and thus is able to find even better partitions of the eigenvalues leading to improved performance.

\begin{table}[H]
\centering
\resizebox{0.6\textwidth}{!}{%

\begin{tabular}{c|cc}
\hline
                               & \textbf{Homophily Rank} & \textbf{Heterophily Rank} \\ \hline
\textbf{LR}                    & 11.25                   & 7.29                      \\
\textbf{MLP}                   & 10.75                   & 5.29                      \\ \hline
\textbf{GCN}                   & 7.88                    & 8.71                      \\
\textbf{SGCN}                  & 6.13                    & 8.57                      \\ \hline
\textbf{SuperGAT}              & 6.25                    & 10.71                     \\
\textbf{H2GCN}                 & 6.88                    & \textbf{4.43 ($3^{rd}$)}       \\
\textbf{FAGCN}                 & 4.38                    & 5.86                      \\
\textbf{GPR-GNN}               & \textbf{2.88 ($1^{st}$)}     & 5.57                      \\
\textbf{APPNP}                 & \textbf{3.50 ($2^{nd}$)}     & 5.86                      \\ \hline
\textbf{EigenNetwork}          & 10.75                   & 9.00                      \\
\textbf{Eigen-EigenNetwork}    & 5.88                    & \textbf{4.29 ($2^{nd}$)}       \\
\textbf{RegEigen-EigenNetwork} & \textbf{4.00 ($3^{rd}$)}     & \textbf{2.00 ($1^{st}$)}       \\ \hline
\end{tabular}
}
\vspace{2mm}
\caption{Average Ranking of Models on Homophilic and Heterophilic datasets.}
\label{tab:rank}
\end{table}

\textbf{Model Comparison: Varying levels of Homophily:} We group all the datasets with homophily score $\leq$ 0.50 and refer to them as heterophilic datasets. The rest datasets are referred to as homophilic datasets. For all the models in comparison, we compute their rank across datasets and report the average rank on homophlilic and heterophilic datasets in Table~\ref{tab:rank}. We used the heterophily datasets result table (Table~\ref{tab:main_table}) and Table~\ref{tab:HomophilyTable} to compute the ranks.

We further group the models for ease of comparison. We observe that the simple non-graph based models, linear and nonlinear networks that use only node features perform reasonably well on heterophilic datasets. The second class of models like SGCN and GCN belong to popular neighborhood aggregation based methods. We see that these models perform better on homophilic datasets, as expected. However, on heterophilic datasets, they perform poor because of the violation of the assumption: connected neighbors have same class labels. 

The third group of models including \supergat, \gprgnn, \fagcn, and \appnp~were specifically designed to work across datasets with varying levels of homophily. We notice that \supergat~performs poorly on heterophilic datasets. We believe that attention trained on auxiliary task alone may not be sufficient to address heterophily. A detailed investigation is required to understand the gaps, and is beyond the scope of this work. However, the other models in this group - \gprgnn, \fagcn~and \appnp~have better average ranking compared to common GNN methods like \gcn~and \sgcn. For homophilic datasets, we observe that models like \appnp~and \gprgnn~perform the best. We noticed in \gprgnn~paper that they perform better than \appnp~on several datasets. However, we see \appnp~has an edge over \gprgnn~in our experiments. We believe this is the case because of amount of labeled data and different splits used in our experiments. In specific, \gprgnn~reports numbers in their paper using 60\% training data for heterophilic datasets, while we follow \cite{geomgcn} and use 48\% training data.

The last set in Table~\ref{tab:rank} corresponds to proposed models. We observe that \eigennet~model does not perform well on several datasets. This is not surprising because \eigennet~ solely relies on topological features for solving the task at hand. However, the merits of this model can be observed on datasets like \textbf{Squirrel} and \textbf{Chameleon}. Therefore, \eigennet~ is still useful in some scenarios. \eigeneigennet~ makes use of node features with neighborhood aggregation and performs significantly better compared to \eigennet~ and other baselines. \regeigeneigenet~offers the best performance on heterophilic datasets and does competitively on homophilic datasets.

\section{Conclusion and Future Work}
\label{sec:conclusion}
In this paper, we presented an eigendecomposition based approach and proposed the \eigennet~models. These models are inspired by the \gprgnn~\cite{gprgnn} model which we show can be interpreted as selecting/weighing the eigenvectors by scaling the corresponding eigenvalues. We propose a weight tying based regularization model, that enables our model to avoid overfitting on the data and generalize better. We show that our models do well across all heterophilic datasets. We plan to study the optimization of $\beta$ variable in \eigeneigennet~model and behaviour of this model on homophily datasets as part of our future work. In this paper, we also propose an alternative concatenation based model that is competitive with aggregation based approach on heterophilic datasets. This model is simple and computationally cheaper. It begs the question whether there are alternative ways to model Graph Neural Networks that work across varying homophily scores. We leave it as a future work.

\newpage
\bibliography{references}
\bibliographystyle{plainnat}

% \input{checklist}
% \clearpage
% \appendix
% \input{sections/appendix}
% \appendix

% \input{sections/appendix}

\end{document}